\documentclass[final,times]{elsarticle}
\usepackage{etex}
\usepackage{empheq}
\usepackage{hyperref}
\usepackage{booktabs,multirow, tabularx} % For formal tables
\usepackage{times,natbib,caption}
\usepackage{bm,array,mathtools,cases,thmtools,thm-restate}
\usepackage{amsthm,amssymb,amsmath}
\usepackage{pgfplots,tikz,graphicx}
\usepackage{subfig}
\usepackage[ruled,linesnumbered]{algorithm2e}

\journal{Elsevier}

\begin{document}
\begin{frontmatter}

\title{Fast single image dehazing via\\multilevel wavelet transform based optimization} 

\author[nus]{Jiaxi He}
\ead{dcshej@nus.edu.sg}
\author[ntu]{Frank Z. Xing}
\ead{zxing001@e.ntu.edu.sg}
\author[sys]{Ran Yang}
\ead{yangran@mail.sysu.edu.cn}
\author[swin]{Cishen Zhang}
\ead{cishenzhang@swin.edu.au}

\address[nus]{Singapore Data Science Consortium, National University of Singapore}
\address[ntu]{School of Computer Science \& Engineering, Nanyang Technological University}
\address[sys]{School of Data and Computer Science, Sun Yat-Sen University}
\address[swin]{School of Software and Electrical Engineering, Swinburne University of Technology}

\begin{abstract} 
The quality of images captured in outdoor environments can be affected by poor weather conditions such as fog, dust, and atmospheric scattering of other particles. This problem can bring extra challenges to high-level computer vision tasks like image segmentation and object detection. However, previous studies on image dehazing suffer from a huge computational workload and corruption of the original image, such as over-saturation and halos. In this paper, we present a novel image dehazing approach based on the optical model for haze images and regularized optimization. Specifically, we convert the non-convex, bilinear problem concerning the unknown haze-free image and light transmission distribution to a convex, linear optimization problem by estimating the atmosphere light constant. Our method is further accelerated by introducing a multilevel Haar wavelet transform. The optimization, instead, is applied to the low frequency sub-band decomposition of the original image. This dimension reduction significantly improves the processing speed of our method and exhibits the potential for real-time applications. Experimental results show that our approach outperforms state-of-the-art dehazing algorithms in terms of both image reconstruction quality and computational efficiency. For implementation details, source code can be publicly accessed via \texttt{http://github.com/JiaxiHe/Image-and-Video-Dehazing}.
\end{abstract}

\end{frontmatter}

%\linenumbers

\newpage
\section{Introduction}
Haze is a result of the accumulation of dust and smoke particles in the atmosphere. It causes degradation of visibility and contrast in images due to the light scattered on haze particles along the transmission path between the object and the sensor. Such a problem prevails in many high-level applications, e.g.,~photography stylization~\cite{rosa17}, semantic segmentation~\cite{xing16}, scene understanding~\cite{christos18}, and video surveillance~\cite{goyal18}, to name a few. The image dehazing problem endeavors to apply post-processing of hazed images to remove the haze effects and reconstruct the original image scene. In recent years, a great deal of progress has been achieved on this crucial topic of image processing~\cite{singh-18}. Wang et al.~\cite{wangw18} categorizes image dehazing methods into the enhancement-based group and the restoration-based group, where the effectiveness of the later is backed up with a sound physical model. However, the optical image model in its original form has several unknown parameters, making single image dehazing an ill-posed problem. A more serious problem is that the haze-free image and the light transmission distribution in atmosphere are bilinearly coupled. As a result, the image restoration process becomes a non-convex problem, which is computationally expensive and easy to converge to a sub-optimal solution.

Early efforts to mitigate this problem often require additional information to reconstruct the image visibility. For instance, Schechner et al. proposed an approach to remove haze by taking two images through one polarizer at different orientations of the same scene~\cite{Schechner2006,Schechner2007}. Hautiere et al.~\cite{Hautiere2007} used the 3D rough geometrical information of the images for analysis; Kopf~\cite{Kopf2008} also used the depth information of the scene. Recently, Luan et al.~\cite{luan17} proposed a learning framework to automatically extract haze-relevant features and estimate light transmission distribution by sampling haze/clear image pairs. The applications of these methods are limited by their requirement of the additional information, which is not always available in practice. For this reason, methods that only use the single hazed image as an input for dehazing have gained increasing attention. To obtain the prior information from the hazed image, He et al.~\cite{He2011} proposed the dark channel prior (DCP) which was later improved in~\cite{Guided}. However, the DCP-based method involved low efficient pixel computation and over-correction when estimating the light transmission distribution. Gao~\cite{Gao2014} observed that contrast and saturation of images can be increased by using the negative image to rectify the hazed image, which is faster than the DCP-based transmission map.

An alternative source of the prior knowledge of haze can be haze scenery modeling. A number of methods~\cite{Fattal2008,Bayesian-defog,Mutimbu,Wang2014} used the Markov random field (MRF) or its variants to estimate the depth information based on a statistical analysis of spatial and contextual dependency of physical phenomena. While some of these methods, e.g.~\cite{Wang2014}, could produce improved results, they generally suffered from additional computational complexity. On the other hand, some MRF-based fast or real-time dehazing methods have the common problem of distorted colors and obvious halos due to their model deficiency. These methods~\cite{Tarel2009,Ancuti2011,Zhang2011,Kim2013,Zhu2015} improved the processing speed at the cost of dehazing quality. Specifically, Kim et al.~\cite{Kim2013} estimated the transmission distribution by maximizing block-wise contrast and at the same time minimizing the information loss due to pixel-level over-saturation. The method was further refined by employing a hierarchical searching technique to decide sky regions. Unfortunately, the detected region can be wrong when bright objects are placed in a close distance. The algorithm of Ancuti et al.~\cite{Ancuti2011} significantly reduced the complexity of DCP by modifying the block-based approach to a pixel-wise one. Although this method has impressively fast processing time, the pixel-wise haze detection is not robust and often suffers from large recognition errors similar to~\cite{Kim2013}. Consequently, the dehazing quality of these two methods is not always visually pleasing. Their time complexity advantages are later surpassed by Zhu et al.~\cite{Zhu2015}, which reported a faster processing speed. Zhao et al.~\cite{Zhao-mof} provide a systematic comparison for the state-of-the-art dehazing methods till present.

We propose a new regularized optimization method termed multilevel wavelet transform based optimization (MWTO) to solve the dehazing problem. Our approach elegantly balances between the image dehazing quality and the processing speed. The formulation of the optimization problem is based on the optical image model, where haze-free image and light transmission distribution are unknown. Inspired by~\cite{Yang2012}, we resolve the non-convex difficulty of the original problem by formulating the bilinearly coupled terms as a whole single term~\cite{he16}.
To further improve the computational efficiency, we perform the discrete Haar wavelet transform (DHWT) on the hazed image to derive a sub-band hazed image model with reduced dimension. Due to the low pass and smoothness characteristic of the light transmission distribution, a piecewise constant assumption on the light transmission distribution is introduced. Based on this assumption, solving the dehazing problem of the sub-band hazed image model with the reduced dimension is sufficient for the solution of the original dehazing problem. This property can significantly reduce the theoretical computational complexity of our method.

Image dehazing with MWTO has several \emph{advantages} compared to its peer methods. \textbf{First}, the formulation of the regularized optimization is a systematic and deterministic approach to a feasible solution for reconstruction of the haze-free image given the optical image model. The regularized optimization problem is computationally efficient and can be readily implemented with standard procedures and software, such as CVX in Matlab and CVXOPT in Python. \textbf{Second}, the regularization terms of the optimization formulation can provide flexibility in computing the dehazing solution by incorporating a priori knowledge of the image and the atmosphere light transmission distribution, which can guide the computation to a meaningful solution. \textbf{Finally}, the proposed sub-band decomposition procedure enables significant dimension reduction. These advantages result in high-quality dehazing performance in very limited time. Experiments suggest that our approach is even faster than linear models, such as in~\cite{Zhu2015}.

The remainder of this paper is organized as follows. Section~\ref{sec:model} provides a brief background of the optical model of haze and its application to the wavelet-decomposed images. Section~\ref{sec:optimization} describes the convex transformation of the sub-band image model and the fusion of multilevel light transmission distribution estimations. In Section~\ref{sec:results}, we evaluate our method and compare to the state-of-the-art methods from many perspectives. Section~\ref{sec:conclusion} concludes the paper with future directions.

\section{Hazed Image Modeling}\label{sec:model}
\subsection{The optical image model}
According to the optical model of haze, the hazed image has two additive components. The first component represents reflected light from the object surface, i.e.~the clear image. The second component is the scattering transmission, i.e.~the haze. We denote the observed digital image in the RGB color space as $\mathbf{I}\in \mathbb{R}^{M\times N\times 3}$ and the haze-free image scene as $\mathbf{J}\in \mathbb{R}^{M\times N\times 3}$, $M=2^i$ and $N=2^j$ with $i$ and $j$ being positive integers, and $m, n \in \Omega$ are indices of the 2-dimensional $M\times N$ index set $\Omega$. Let $c=1,2,3$ be the color channel index, then $\mathbf{I}_c, \mathbf{J}_c\in \mathbb{R}_+^{M\times N}$ are matrices with non-negative entries denoted by $I_c(m,n)$ and $J_c(m,n)$, respectively. The optical model can be written as:
\begin{equation}
\label{eq:model}
\mathbf{I}_c=\mathbf{J}_c\odot\mathbf{t}+a_c (\mathbf{1}-\mathbf{t}),\ \ \ \ c=1,2,3,
\end{equation}
where each $a_c$ is the atmospheric light constant of the corresponding color channel, $\mathbf{t} \in \mathbb{R}_+^{M\times N}$ is the transmission distribution representing the portion of the light, not being scattered,  illuminating on camera sensors, $\odot$ denotes the elementwise multiplication operation~\footnote{In the remainder of this paper, we follow this denotation and use the symbols $\oplus$, $\ominus$, and $\oslash$ to represent elementwise addition, subtraction, and division respectively.} and $\mathbf{1}$ is the matrix of appropriate dimension with all-one entries. The hazy scene image $\mathbf{I}_c$ is the result of the attenuated image intensity $\mathbf{J}_c$ through the scattering transmission path, together with the scattered transmission atmospheric light.
In the following discussion, we assume the entries of $\mathbf{I}_c$, $\mathbf{J}_c$, $a_c$ and $\mathbf{t}$ are unitized, such that
$\mathbf{0}\preceq \mathbf{I}_c \preceq \mathbf{1}$, $\mathbf{0}\preceq \mathbf{J}_c \preceq \mathbf{1}$, $0 < a_c \leq 1$, $c=1,2,3$, and $\mathbf{0} \prec \mathbf{t} \preceq \mathbf{1}$, where $\mathbf{0}$ is a matrix with all-zero entries and $\preceq$ ($\prec$) is the elementwise operation of $\leq$ ($<$) on matrices.

In practice, $\mathbf{I}_c$ is the only observable image and $\mathbf{J}_c$, $a_c$, $\mathbf{t}$ are unknown. The objective of image dehazing is to
estimate $\mathbf{J}_c$, as well as $a_c$ and $\mathbf{t}$, so to reconstruct the composed haze free color image $\mathbf{J}= \mathbf{J}_1\oplus \mathbf{J}_2\oplus \mathbf{J}_3$.

\subsection{The sub-band image model} \label{sec:subb}
\begin{figure}[htbp]
\centering
\includegraphics[width=\textwidth]{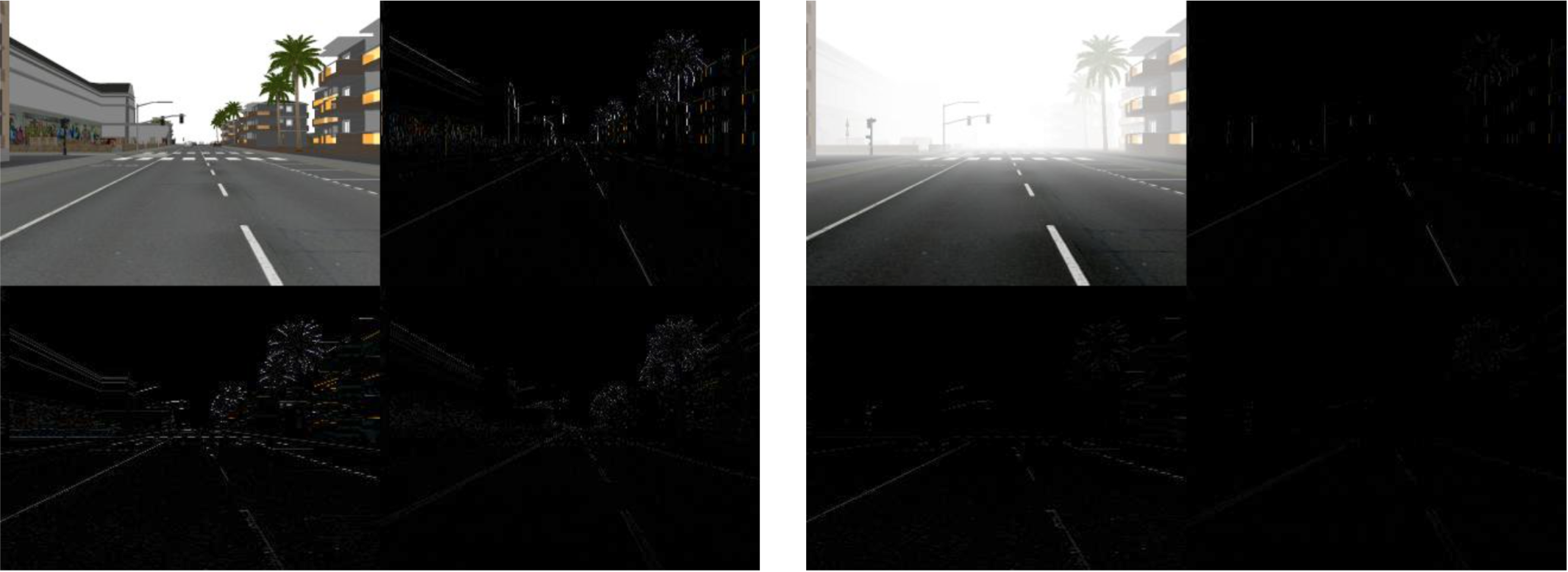}
\caption{Single-level DHWT results for a clear image and the corresponding simulated hazed image, performed on the FRIDA dataset of~\cite{Tarel2012}.}
\label{flh}
\end{figure}

Given the assumption by the optical image model that the transmission rate is evenly distributed in atmosphere, the frequency response of haze in images should be mainly distributed within the low-frequency sub-band. This hypothesis is testified in Fig.~\ref{flh}. Based on this low pass characteristic of the light transmission distribution, we apply DHWT to fast decompose the image model into a bank of frequency sub-bands. The decomposed image model in the low-frequency band thus already contains the information of light transmission distribution $\mathbf{t}$ and can be used for its estimation. This sub-band image model with reduced dimension can result in significant reduction of the computational complexity in the dehazing process.

Specifically, when the atmosphere is homogeneous, the light transmission distribution $\mathbf{t}$ can be represented by
\begin{equation}\label{eq:t(x)}
\mathbf{t}=e^{-\beta \mathbf{d}},
\end{equation}
where $\mathbf{d} \in \mathbb{R}^{M\times N}$ is the distance map from the target object to the camera, $\beta$ is the scattering coefficient depending on the hazy medium. Both $\mathbf{d}$ and $\beta$ are strictly positive, which implies that $\mathbf{t}$ is elementwise bounded by the all-zero and all-one matrices, i.e.~$\mathbf{0} \prec \mathbf{t} \preceq \mathbf{1}$. Sampled from the physical world, pixels in each geometric local patch share approximately the same depth value to constitute a region or object. Abrupt depth jump of pixel values, in contrast, constitute edges of objects or regional boundaries. Therefore, it is evident to assume that the distance map $\mathbf{d}$ is piecewise constant for most images. Since $\mathbf{t}$ is a continuous map of $\mathbf{d}$ in (\ref{eq:t(x)}), $\mathbf{t}$ is also piecewise constant.
With this consideration, it is assumed that the $M\times N$ dimensional distribution $\mathbf{t}$ is $2$-patch piecewise constant in the sense that $\mathbf{t}(2m+i,2n+j)=\mathbf{t}(2m,2n)$, for $m\in[0,M/2)$, $n\in[0,N/2)$ and $i,j=0,1$. Using this assumption and the DHWT, the sub-band image model can be further specified as following.

Let $\mathbf{W}$ be the 2-dimensional DHWT matrix of appropriate dimension, that is, 
\begin{empheq}[left=\mathbf{W}_{p, q}^n \doteq \empheqlbrace]{align}
&1/\sqrt{2^{k-p}} \quad when \quad q\times 2^{k-p} \leq n < (q+\frac{1}{2})\times 2^{k-p} \\
&-1/\sqrt{2^{k-p}} \quad when \quad (q+\frac{1}{2})\times 2^{k-p} \leq n < (q+1)\times 2^{k-p} \\
&0  \quad otherwise.
\end{empheq}

Subsequently, the single-level DHWT of $\mathbf{I}_c$ and $\mathbf{J}_c$, $c=1,2,3$, gives results in their transformed matrices with four ${M\over 2}\times {N\over 2}$ dimensional sub-band blocks, i.e.,
\begin{align}
\mathbf{\hat{I}}_c &= \mathbf{W} \mathbf{I}_c \mathbf{W}^T
=\small\left[\begin{array}{ll}
\mathbf{\hat I}_c^a & \mathbf{\hat I}_c^h \\
\mathbf{\hat I}_c^v & \mathbf{\hat I}_c^d\\
\end{array}\right], \label{w1}\\
\mathbf{\hat{J}}_c &= \mathbf{W} \mathbf{J}_c \mathbf{W}^T
=\small\left[\begin{array}{ll} \mathbf{\hat J}_c^a & \mathbf{\hat J}_c^h \\
\mathbf{\hat J}_c^v & \mathbf{\hat J}_c^d\\
\end{array}\right], \label{w2}
\end{align}
where the superscripts $a$, $h$, $v$ and $d$ indicate the low-frequency approximation, horizontal, vertical, and diagonal sub-band blocks of the wavelet transform respectively. If the light transmission distribution $\mathbf{t}$ is $2$-patch piecewise constant, it can be verified that its single-level DHWT is
\begin{align}\label{eq:t}
\mathbf{\hat{t}}= \mathbf{W} \mathbf{t} \mathbf{W}^T
=\small\left[\begin{array}{ll}
2\mathbf{\hat t}^a & \mathbf{0} \\
\mathbf{0} & \mathbf{0}\\
\end{array}\right],
\end{align}
where $\mathbf{\hat t}^a$ is the low pass sub-band distribution of the DHWT of $\mathbf{t}$ satisfying
$${t}^a(m,n)=t(2m,2n),\ \ \ \ m,n \in \Omega^a,$$
with $\Omega^a$ being the 2-dimensional index set of the low pass sub-band distribution $\mathbf{\hat t}^a$.

Using (\ref{w1}), (\ref{w2}), and (\ref{eq:t}), we derive the DHWT-based optical image model as~(\ref{eq:DHWTmodel}):
\begin{equation}
\label{eq:DHWTmodel}
{\small \left[\begin{array}{cc} \mathbf{\hat I}_c^a & \mathbf{\hat I}_c^h \\
\mathbf{\hat I}_c^v & \mathbf{\hat I}_c^d\\
\end{array}\right]=\left[\begin{array}{cc} \mathbf{\hat J}_c^a \odot \mathbf{\hat t}^a + 2a_c(\mathbf{1}-\mathbf{\hat t}^a)& \mathbf{\hat J}_c^h \odot \mathbf{\hat t}^a \\
\mathbf{\hat J}_c^v\odot \mathbf{\hat t}^a  & \mathbf{\hat J}_c^d \odot \mathbf{\hat t}^a\\
\end{array}\right]}.
\end{equation}
where the low pass sub-band block of the matrix equation (\ref{eq:DHWTmodel}) presents a DHWT sub-band image model with a reduced dimension of ${M\over 2}\times {N\over 2}$ as follows
\begin{equation}
\label{eq:DHWTmodel2}
\mathbf{\hat I}_c^a=\mathbf{\hat J}_c^a \odot \mathbf{\hat t}^a + \hat a_c(\mathbf{1}-\mathbf{\hat t}^a), \ \ \ \ \hat a_c=2a_c, \ c=1,2,3.
\end{equation}
Note that this low pass sub-band image model with reduced dimension has the exact same form as that of the original optical image model in (\ref{eq:model}), except for that the airlight $\hat a_c$ doubles the color shift effect in the low-frequency sub-band.
Meanwhile, the high-frequency sub-band blocks has the following hazed model:
\begin{align}
\label{eq:approximation2}
&\hat{\mathbf{I}}_c^h=\hat{\mathbf{J}}_c^h\odot \mathbf{t}_d\\
&\hat{\mathbf{I}}_c^v=\hat{\mathbf{J}}_c^v\odot \mathbf{t}_d\\
&\hat{\mathbf{I}}_c^d=\hat{\mathbf{J}}_c^d\odot \mathbf{t}_d
\end{align}
where  the high-frequency coefficients are free from color shift and are only weakened by the down-sampled transmission function $\mathbf{t}_d$. Consequently, instead of recovering $\mathbf{\hat J}_c$, the sub-band image model only requires recovering $\mathbf{\hat J}_c^a$ and $\mathbf{t}_d$, such that the high-frequency coefficients can be easily derived by:
\begin{equation}
\hat{\mathbf{J}}_c^h=\hat{\mathbf{I}}_c^h\oslash\mathbf{t}_d, \ \ \ \hat{\mathbf{J}}_c^v=\hat{\mathbf{I}}_c^v\oslash\mathbf{t}_d, \ \ \
\hat{\mathbf{J}}_c^d=\hat{\mathbf{I}}_c^d\oslash\mathbf{t}_d.
\end{equation}
where $\oslash$ denotes the elementwise division operation on matrices.

Finally, $\mathbf{J}_c$ can be reconstructed using inverse discrete Haar wavelet transform. Paring equations~(\ref{eq:model}) and~(\ref{eq:DHWTmodel}), we have the general sub-band hazed image model when multiple levels of wavelet decomposition are recursively performed:
\begin{align}
\label{eq:model2}
\hat{\mathbf{I}}_c^a&=\hat{\mathbf{J}}_c^a\odot \mathbf{t}_d^l+2^{l}a_c(\mathbf{1}-\mathbf{t}_d^l)\nonumber\\
&=\hat{\mathbf{J}}_c^a\odot \mathbf{t}_d^l+\hat{a}_c^a(\mathbf{1}-\mathbf{t}_d^l), \ \ \ c=1,2,3,
\end{align}
where $l=0,1,2,...$ is the level of wavelet decomposition, $\hat{a}_c^a=2^la_c$ is the low-frequency sub-band's airlight and $\mathbf{t}_d^l$ is the sub-band's light transmission distribution that has the form of
\begin{equation}
\mathbf{t}_d^l(m,n)=\mathbf{t}(2^lm,2^ln).
\end{equation}
Subsequently, the dehazing algorithm is only performed in the low-frequency sub-band block, and the high-frequency coefficients are recovered by dividing the corresponding $\mathbf{t}_d$.

\section{Image Dehazing via Optimization}\label{sec:optimization}
\subsection{Atmospheric light estimation}
Estimating the atmospheric light constants $a_c$, $c=1,2,3$ is an important starting point for image dehazing. Many previous studies, e.g.~\cite{Narasimhan2002,Tarel2009,He2011} estimated $a_c$ with the most haze-opaque region, though it may be affected by white objects in the scene~\cite{wangw17}. Others attempted to find this region by sophisticated techniques, e.g. hierarchical searching~\cite{Kim2013}. However, as aforementioned, the robustness of these techniques are questionable. For simplicity, we consider the brightest pixel as the estimates of $a_c$ through filtering~\cite{airlight}, i.e.
\begin{equation}
\label{eq:ac}
\hat a_c=\max_{{m,n}\in \Omega} \ \min_{{k,l}\in \omega(m,n)} I_c(k,l), \ \ \ \ c= 1,2,3,
\end{equation}
where  $\omega(m,n)$ is a $3\times 3$ local window centered at $(m,n)$.
Using this result, we assume that the estimates of $a_c$, $c=1,2,3$, have been obtained and are used in the hazed image models for further estimation of $\mathbf{J}_c$ and $\mathbf{t}$.

\subsection{Linear formulation of the sub-band image model}
Recall the low dimensional DHWT-based sub-band model in (\ref{eq:DHWTmodel2}), where the low sub-band blocks $\mathbf{\hat I}_c^a$ and the estimated airlight ${\hat a}_c=2a_c$, $c=1,2,3$ are known. Joint estimation of $\mathbf{\hat J}_c^a$ and $\hat{\mathbf{t}}^a$ is still a bilinearly coupled problem. However, we observe that if we consider $\mathbf{\hat J}_c^a \odot \mathbf{\hat t}^a$ as a whole, the sub-band hazed image model can be converted to a convex, linear optimization problem. For this reason we introduce the substitution:
$$\mathbf{\hat Q}_c^a=\mathbf{\hat J}_c^a \odot \mathbf{\hat t}^a, \ c=1,2,3,$$
$$\mathbf{\hat Y}_c^a=\mathbf{\hat I}_c^a - \hat a_c \mathbf{1}, \ c=1,2,3.$$
Then, the sub-band image model (\ref{eq:DHWTmodel2}) can be written as
\begin{equation}
\label{eq:DHWTmodel3}
\mathbf{\hat Y}_c^a=\mathbf{\hat Q}_c^a - \hat a_c \mathbf{\hat t}^a, \ c=1,2,3,
\end{equation}
where variable $\mathbf{\hat Y}_c^a$ and the estimated airlight $\hat a_c$ are known. The image dehazing problem is thus formulated as solving $\mathbf{\hat Q}_c^a$ and $\mathbf{\hat t}^a$ from the new sub-band model (\ref{eq:DHWTmodel3}). The solutions for $\mathbf{\hat Q}_c^a$ and $\mathbf{\hat t}^a$ are sufficient for further estimation of the wavelet transformed sub-band image blocks. To elaborate, given $\mathbf{\hat Q}_c^a$ and $\mathbf{\hat t}^a$, it follows from (\ref{eq:DHWTmodel}), $\mathbf{\hat Q}_c^a=\mathbf{\hat J}_c^a \odot \mathbf{\hat t}^a$, and $\mathbf{0} \prec \mathbf{\hat t}^a$, that the wavelet transformed sub-band image blocks can be estimated by
\begin{equation}\label{subbandblocks} \begin{array}{lll}
\mathbf{J}_c^{a}=\mathbf{Q}_c^a \oslash\mathbf{\hat t}^a, &
\mathbf{J}_c^{h}=\mathbf{I}_c^h \oslash\mathbf{\hat t}^a, & \\
\mathbf{J}_c^{v}=\mathbf{I}_c^v \oslash\mathbf{\hat t}^a, &
\mathbf{J}_c^{d}=\mathbf{I}_c^d \oslash\mathbf{\hat t}^a, & c=1,2,3.
\end{array}
\end{equation}
The reconstruction of the haze-free image matrices $\mathbf{J}_c$ can be further obtained by the inverse DHWT of $\mathbf{\hat J}_c$ providing equation (\ref{w2}) and (\ref{subbandblocks}).

\subsection{Regularized optimization for dehazing}
Note that although the linear formulation of the sub-band image model guarantees closed solutions to $\mathbf{\hat Q}_c^a$ and $\mathbf{\hat t}^a$ when $\mathbf{\hat Y}_c^a$ is determined, there are an infinite number of solutions in a continuous space. In order to find the feasible and meaningful solutions for $\mathbf{\hat Q}_c^a$ and $\mathbf{\hat t}^a$, incorporation of available knowledge and information about the image and haze conditions into appropriate constraints on $\mathbf{\hat Q}_c^a$ and $\mathbf{\hat t}^a$ is an indispensable step. The constraints can regulate solutions $\mathbf{\hat Q}_c^a$ and $\mathbf{\hat t}^a$ to satisfactory values. Moreover, the original problem is not compromised since linear transformation (\ref{eq:DHWTmodel3}) guarantees an optimal $(\mathbf{J}_c^{a*},\mathbf{t}_d^*)$ when $(\mathbf{Q}_c^{a*},\mathbf{t}_d^*)$ is an optimal solution to (\ref{eq:DHWTmodel3}).

Based on the sub-band image model (\ref{eq:DHWTmodel3}) which is linear in $\mathbf{\hat t}^a$ and $\mathbf{\hat Q}_c^a$, the general formulation of the proposed regularized convex optimization for dehazing is written as.
\begin{eqnarray}
\label{eq:DHWTmodel4}
\min_{\mathbf{\hat Q}_c^a,\mathbf{t}^a}&&
R(\mathbf{\hat t}^a,\mathbf{\hat Q}_c^a, \ c=1,2,3), \label{opt1} \\
\textrm{s.t.} && \mathbf{\hat Y}_c^a - \mathbf{\hat Q}_c^a + \hat a_c \mathbf{\hat t}^a = \mathbf{0}, \nonumber \\
&& \mathbf{0}\prec \mathbf{\hat t}^a \preceq \mathbf{1}, \ \ \mathbf{0} \preceq \mathbf{\hat Q}_c^a, \ \ c=1,2,3, \nonumber
\end{eqnarray}
where $R(\mathbf{\hat t}^a,\mathbf{\hat Q}_c^a, \ c=1,2,3)$ denotes a convex regularization function of $\mathbf{\hat Q}_c^a$ and $\mathbf{\hat t}^a$ to be selected.

A na\"{i}ve selection of the regularization function can be the mean squared contrast function of $\mathbf{\hat J}_c^a$ given by \cite{Kim2013}:
\begin{eqnarray}
\label{eq:cmse}
C_{ms}&=&\sum_{c=1,2,3; \ {(m,n)}\in \Omega^{a}}
\frac{(\hat J_c^a(m,n)-\bar{J}_c^a)^2}{{N_{\Omega^a}}}\nonumber \\
&=& \sum_{c=1,2,3;\ {(m,n)}\in \Omega^{a}}
\frac{(\hat I_c^a(m,n)-\bar{I}_c^a)^2}{{{\hat t}^a}(m,n)^2\cdot N_{\Omega^a}},
\end{eqnarray}
where $\Omega_a$ denotes the domain of the 2-dimensional pixel index in the low-frequency sub-band block,
$\bar{J}_c^a$, $\bar{I}_c^a$ are the average pixel values of $\mathbf{\hat J}_c^a$ and $\mathbf{\hat I}_c^a$, respectively, and $N_{\Omega^a}$ is the total pixel number. Since the haze effect reduces the degree of contrast in images, the general idea of image dehazing process is to enhance the level of image contrast. A higher value of $C_{ms}$ indicates a higher contrast of the image. The above equation (\ref{eq:cmse}) implies that the image's contrast $C_{ms}$ is inversely proportional to the square of the transmission function $\hat t^a(m,n)$. Therefore, reducing the value of $\hat t^a(m,n)^2$ can improve the contrast of the image. For implementing this consideration, a $\|\mathbf{\hat t}^a\|_F^2$ term is introduced into the regularization function to penalize the values of $\hat t^a(m,n)^2$, where $\|\cdot\|_F$ denotes the Frobenius norm~\cite{higham2002}.

However, the sub-band image model (\ref{eq:DHWTmodel3}) used for the proposed regularized optimization is
primarily an elementwise equation of image pixels. Straightforward elementwise operations based on this image model may not well represent and reconstruct dependency and connectivity information of image pixels with their adjacent neighborhoods. Therefore, it is important and necessary that the regularization function takes into account of dependency and connectivity properties of pixels. Like described in Section~\ref{sec:subb}, the depth map $\mathbf{d}$ is piecewise smooth, which leads to the piecewise constant characteristics of $\mathbf{t}$. Moreover, the down-sampled $\mathbf{t}_d$ also inherits the piecewise constant characteristics. Within a reasonable range, such characteristics can be further extended to its transformed sub-band distribution $\mathbf{\hat t}^a$.
To promote the low pass and piecewise constant characteristics of $\mathbf{\hat t}^a$, a well-known total variation function term $\|\mathbf{\hat t}^a\|_{TV}$~\cite{TV} is introduced into the regularization function, where $\|\cdot\|_{TV}$ denotes the total variation norm.

With the above considerations, we specify the regularization function of our proposed method as:
\begin{eqnarray}
R(\mathbf{t}^a,\mathbf{\hat Q}_c^a,c=1,2,3) = \|\mathbf{\hat t}^a\|_F^2+\lambda \|\mathbf{\hat t}^a\|_{TV},\nonumber
\end{eqnarray}
where $\lambda>0$ is a weighting parameter, balancing the penalty weights on regularization terms to guide the optimization solution to satisfactory values. A guideline to the value of $\lambda$ is to set it small when the haze is thick, emphasizing on contrast enhancement. When the haze is thin, $\lambda$ should be relatively larger.
As a result, the regularized convex optimization for image dehazing is formulated as:
\begin{eqnarray}
\label{opt2}
\min_{\mathbf{\hat t}^a,\mathbf{\hat Q}_c^a }&& \|\mathbf{\hat t}^a\|_F^2 +\lambda \|\mathbf{\hat t}^a\|_{TV}  \\ \nonumber \\
\textrm{s.t.} && \mathbf{\hat Y}_c^a - \mathbf{\hat Q}_c^a + \hat a_c \mathbf{\hat t}^a = \mathbf{0}, \nonumber \\
&& \mathbf{0}\prec \mathbf{\hat t}^a \preceq \mathbf{1}, \ \ \mathbf{0} \preceq \mathbf{\hat Q}_c^a, \ \ c=1,2,3. \nonumber
\end{eqnarray}
Model (\ref{opt2}) is called single-level wavelet transform based optimization (SWTO), which is the essential ingredient of its multilevel extension.

\subsection{Extension to multilevel sub-band models}
\begin{figure}[htbp]
\centering
\includegraphics[width = 0.8\textwidth]{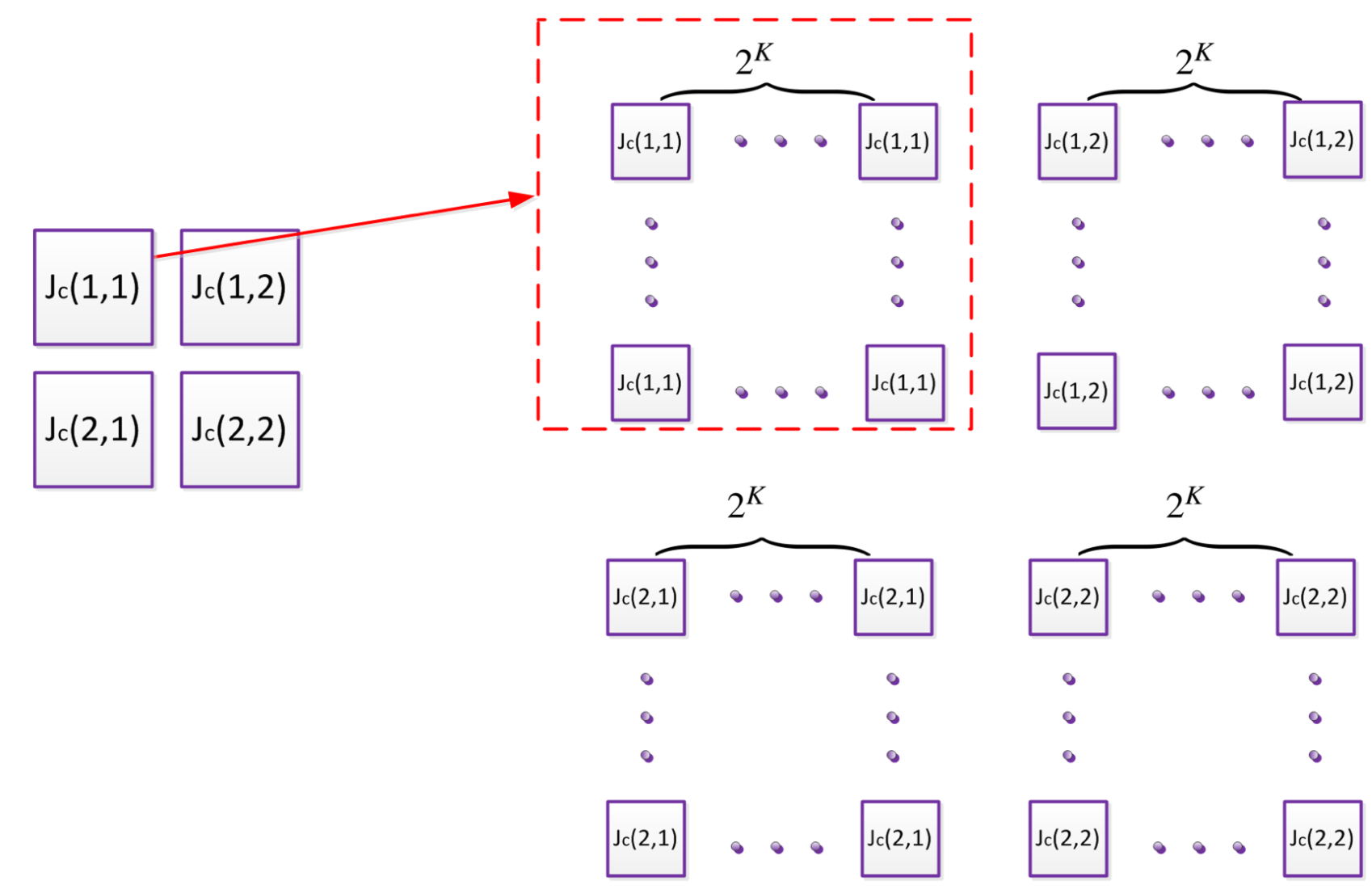}
\caption{Recursively resizing $\mathbf{\hat t}_k$ to obtain the lower-level transmission distribution $\mathbf{\hat t}_{k-1}$.} \label{solvetk}
\end{figure}

Obviously, the conditions for the aforementioned SWTO model are directly applicable to the original hazed image model (\ref{eq:model}) and the sub-band image models with higher-level wavelet decomposition. With an extension of the $2$-patch piecewise constant assumption on the light transmission distribution $\mathbf{t}$ to the $2^k$-patch piecewise constant assumption for $k\geq 2$, the multilevel sub-band model provides further reduced model dimension and computational complexity. Let $\mathbf{W}_k$ be the $k$-th level DHWT matrix, we have:
$$\mathbf{\hat I}_{c,k} =  \mathbf{W}_k\cdots\mathbf{W}_2\mathbf{W}_1\mathbf{I}_c\mathbf{W}_1^T\mathbf{W}_2^T\cdots\mathbf{W}_k^T,$$
$$\mathbf{\hat J}_{c,k} =  \mathbf{W}_k\cdots\mathbf{W}_2\mathbf{W}_1\mathbf{J}_c\mathbf{W}_1^T\mathbf{W}_2^T\cdots\mathbf{W}_k^T.$$
Similarly to the single-level DHWT-based optical image model~(\ref{eq:DHWTmodel}), we derive the multilevel DHWT-based optical image model in an iterated function form as:
\begin{equation}
\label{eq:MWTOmodel}
{\small \left[\begin{array}{cc} \mathbf{\hat I}_{c,k}^a & \mathbf{\hat I}_{c,k}^h \\
\mathbf{\hat I}_{c,k}^v & \mathbf{\hat I}_{c,k}^d\\
\end{array}\right]
=\left[\begin{array}{cc} \mathbf{\hat J}_{c,k}^a \odot \mathbf{\hat t}_k + 2a_{c,k}(\mathbf{1}-\mathbf{\hat t}_k)& \mathbf{B} \\
\mathbf{C} & \mathbf{D}\\
\end{array}\right]}.
\end{equation}
where component matrices $\mathbf{B}=\mathbf{G}_k^M\mathbf{\hat J}_{c,k-1}^a(\mathbf{H}_k^N)^T \odot \mathbf{\hat t}_k$, $\mathbf{C}=\mathbf{H}_k^M\mathbf{\hat J}_{c,k-1}^a(\mathbf{G}_k^N)^T\odot \mathbf{\hat t}_k$, $\mathbf{D}=\mathbf{G}_k^M\mathbf{\hat J}_{c,k-1}^a(\mathbf{G}_k^N)^T \odot \mathbf{\hat t}_k$, $k\geq2$, $\mathbf{H}$ and $\mathbf{G}$ are the low pass averaging Haar transform matrix and the high pass difference Haar transform matrix. The process to recursively solve $\mathbf{\hat t}_k$ and the original transmission distribution $\mathbf{t}$ from model~(\ref{eq:MWTOmodel}) is illustrated with Fig.~\ref{solvetk}.

\section{Experiments and Results}\label{sec:results}
We conduct extensive experiments on a large collection of hazed images and simulated hazed images to evaluate the performance of our proposed MWTO algorithm. Among these, the famous ``Canyon", ``Desk", ``Hill", ``House", ``Lily", ``Mountain", and ``River" images are presented. Moreover, we compare with literature~\cite{Kim2013} on more images from its supplementary materials to show the quality improvement by adding a new regularization term to the conventional function.
We implemented the state-of-the-art image dehazing methods, including Fattal's method~\cite{Fattal2008}, Tarel's method~\cite{Tarel2009}, He's method~\cite{Guided}, Meng's method~\cite{Meng}, Kim's method~\cite{Kim2013}, Wang's method~\cite{Wang2014}, Nishino's method~\cite{Bayesian-defog}, Zhu's method~\cite{Zhu2015}, and Berman's method~\cite{Dana2016}. Among these, the computations of the algorithms of Tarel et al., He et al., Meng et al., Zhu et al., and Berman et al. use the Matlab codes provided by the authors. Performance comparisons with the other algorithms use the results presented in the corresponding publications and authors' websites. Implementation of the MWTO method uses the regularized convex optimization software of the Split Bregman iteration algorithm~\cite{SplitBregman}. All of the algorithms were executed on an HP-Z420 workstation with a 3.30 GHz Intel E5-1660 CPU and parallel computing disabled. We set the multilevel parameter to $2$ for simplicity, and weighting parameter $\lambda$ is set according to $\hat a_c$.

The comparison involves many perspectives, such as subjective evaluation, objective evaluation, and computational complexity analysis. More specifically, subjective evaluation is based on both simulated haze and natural haze datasets~\cite{ihaze,ohaze}; objective evaluation consists of quantitative visibility assessment~\cite{Tarel2008}, mean square evaluation (MSE)~\cite{zwang04}, peak signal-to-noise ratio (PSNR), and structural similarity (SSIM)~\cite{wangw17,mngpp}; computational complexity analysis consists of  comparison of running time and theoretical analysis of our model scalability.

\subsection{Subjective evaluation}
\begin{figure*}
\begin{center}
\subfloat[Hazy Input]{\includegraphics[width=.31\textwidth,height=3cm]{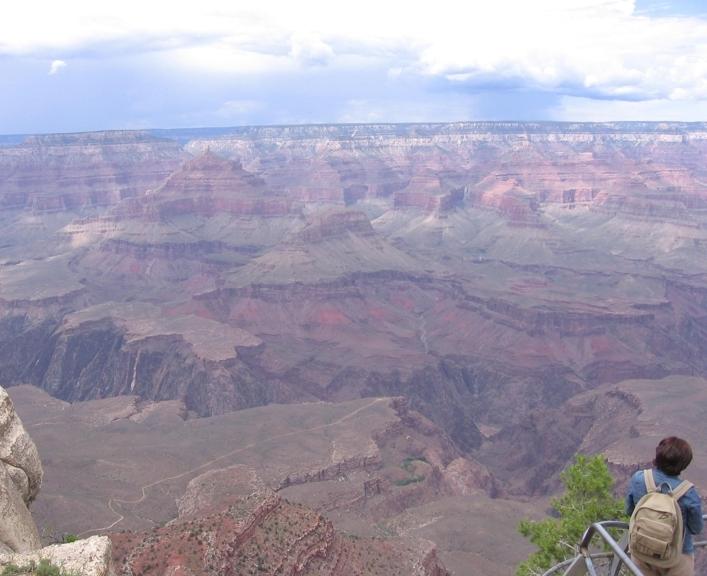}}\quad
\subfloat[Tarel's method~\cite{Tarel2009}]{\includegraphics[width=.31\textwidth,height=3cm]{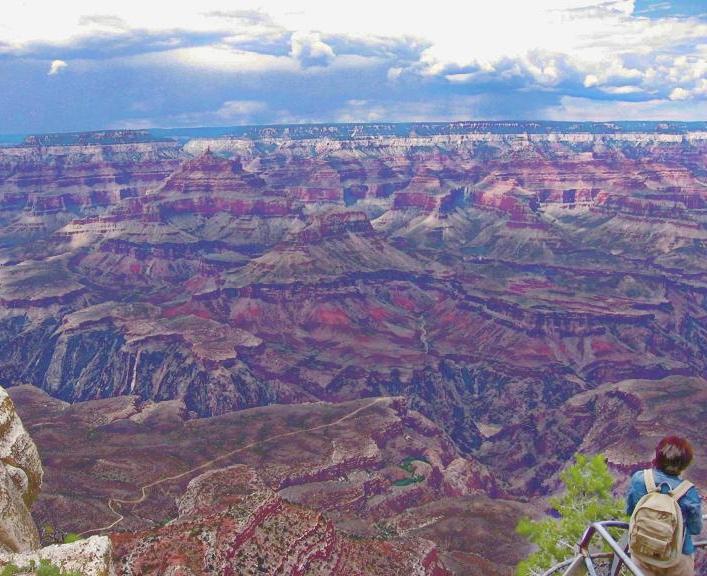}}\quad
\subfloat[He's method~\cite{Guided}]{\includegraphics[width=.31\textwidth,height=3cm]{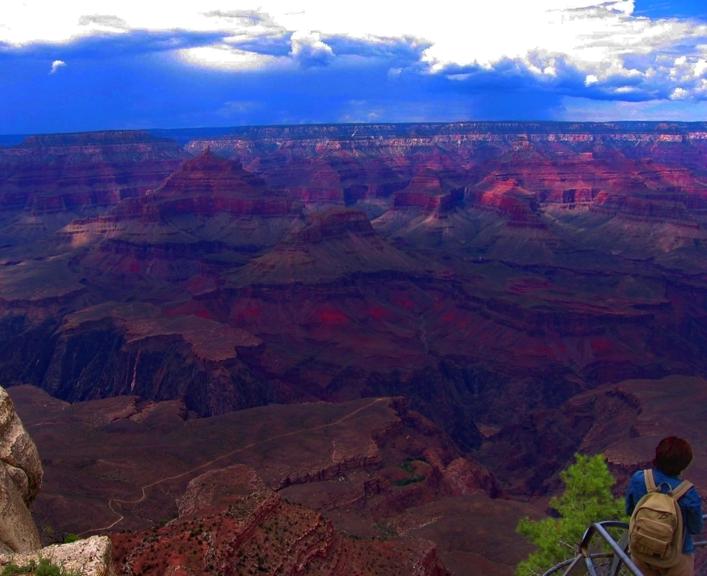}}\quad
\subfloat[Meng's method~\cite{Meng}]{\includegraphics[width=.31\textwidth,height=3cm]{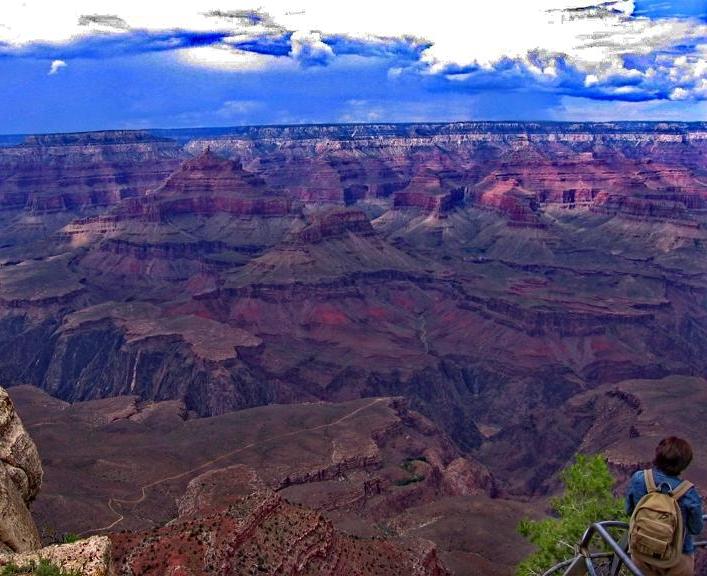}}\quad
\subfloat[Berman's method~\cite{Dana2016}]{\includegraphics[width=.31\textwidth,height=3cm]{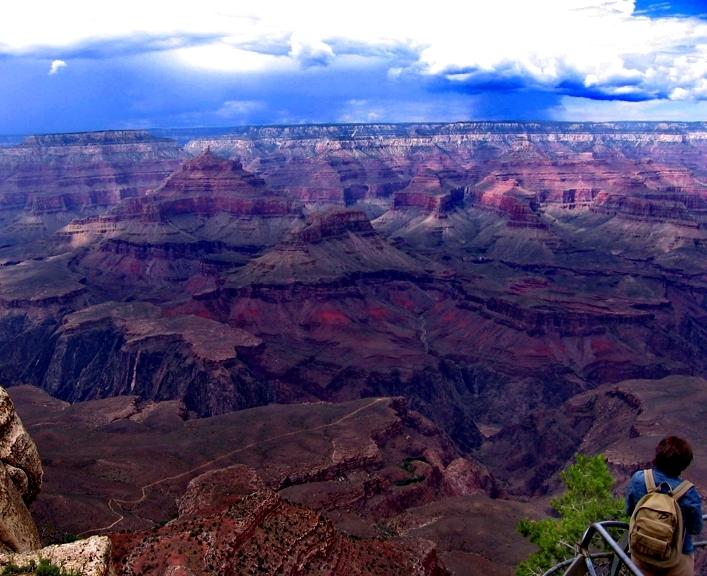}}\quad
\subfloat[Proposed method (MWTO)]{\includegraphics[width=.31\textwidth,height=3cm]{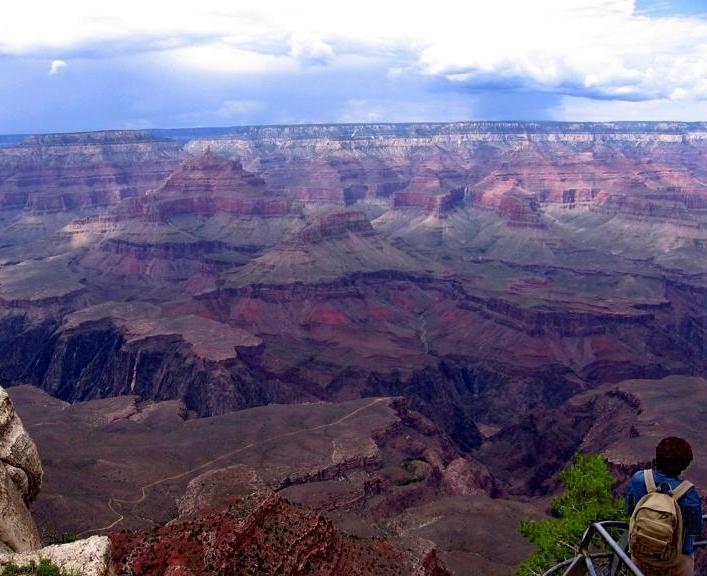}}\\
\caption{Dehazing results by different algorithms for the ``Canyon" image.}\label{fig:Canyon}
\end{center}
\end{figure*}

\begin{figure*}
\begin{center}
\subfloat[Hazy Input]{\includegraphics[width=.31\textwidth,height=3cm]{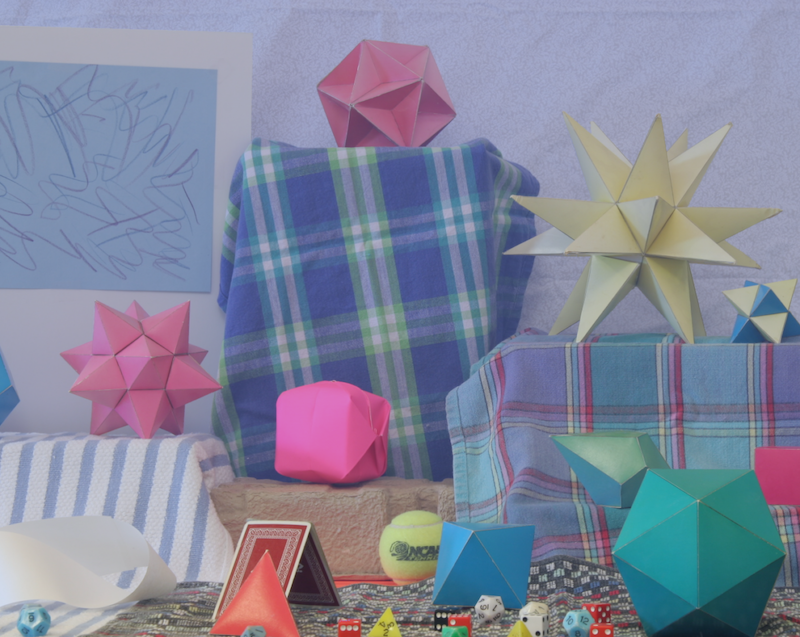}}\quad
\subfloat[Zhu's method~\cite{Zhu2015}]{\includegraphics[width=.31\textwidth,height=3cm]{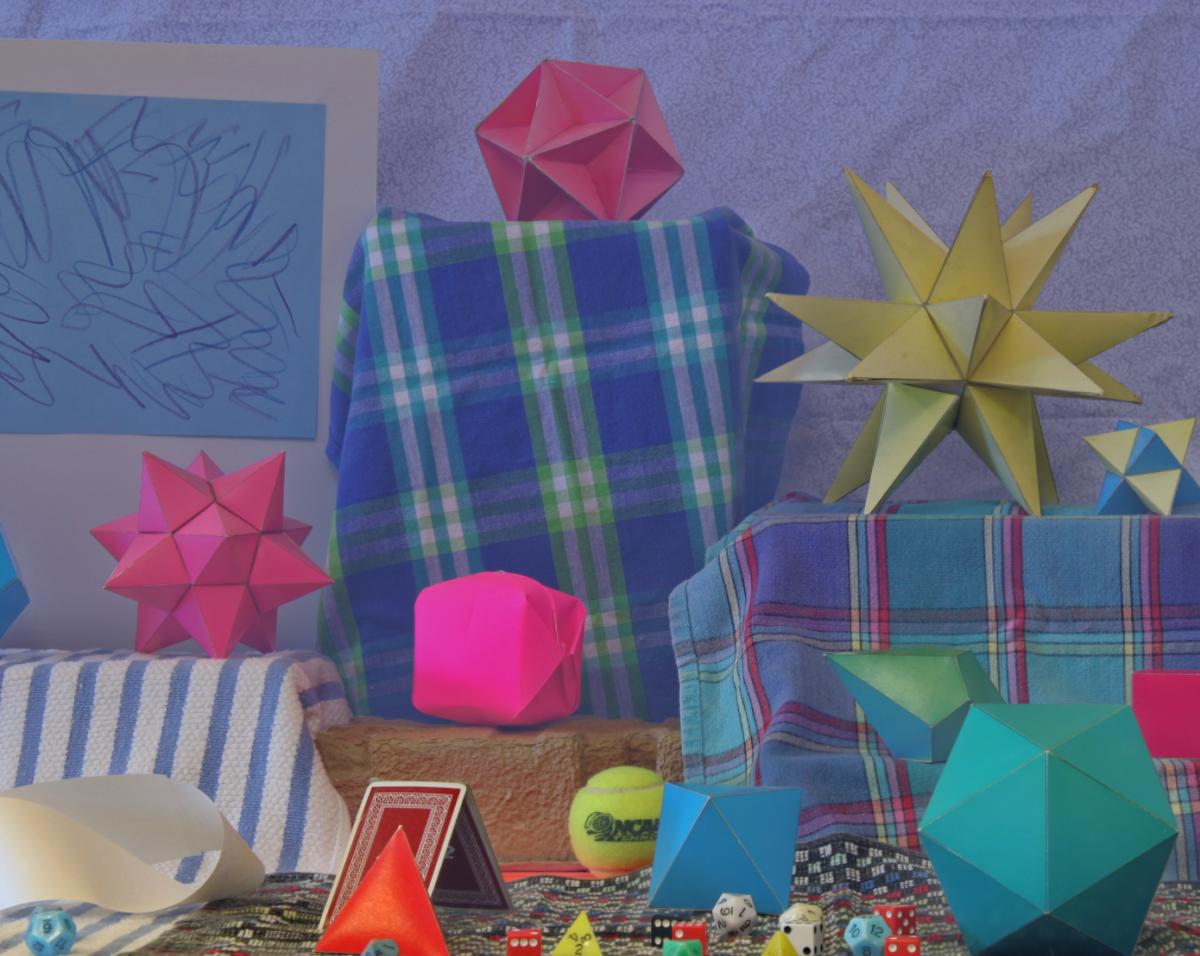}}\quad
\subfloat[He's method~\cite{Guided}]{\includegraphics[width=.31\textwidth,height=3cm]{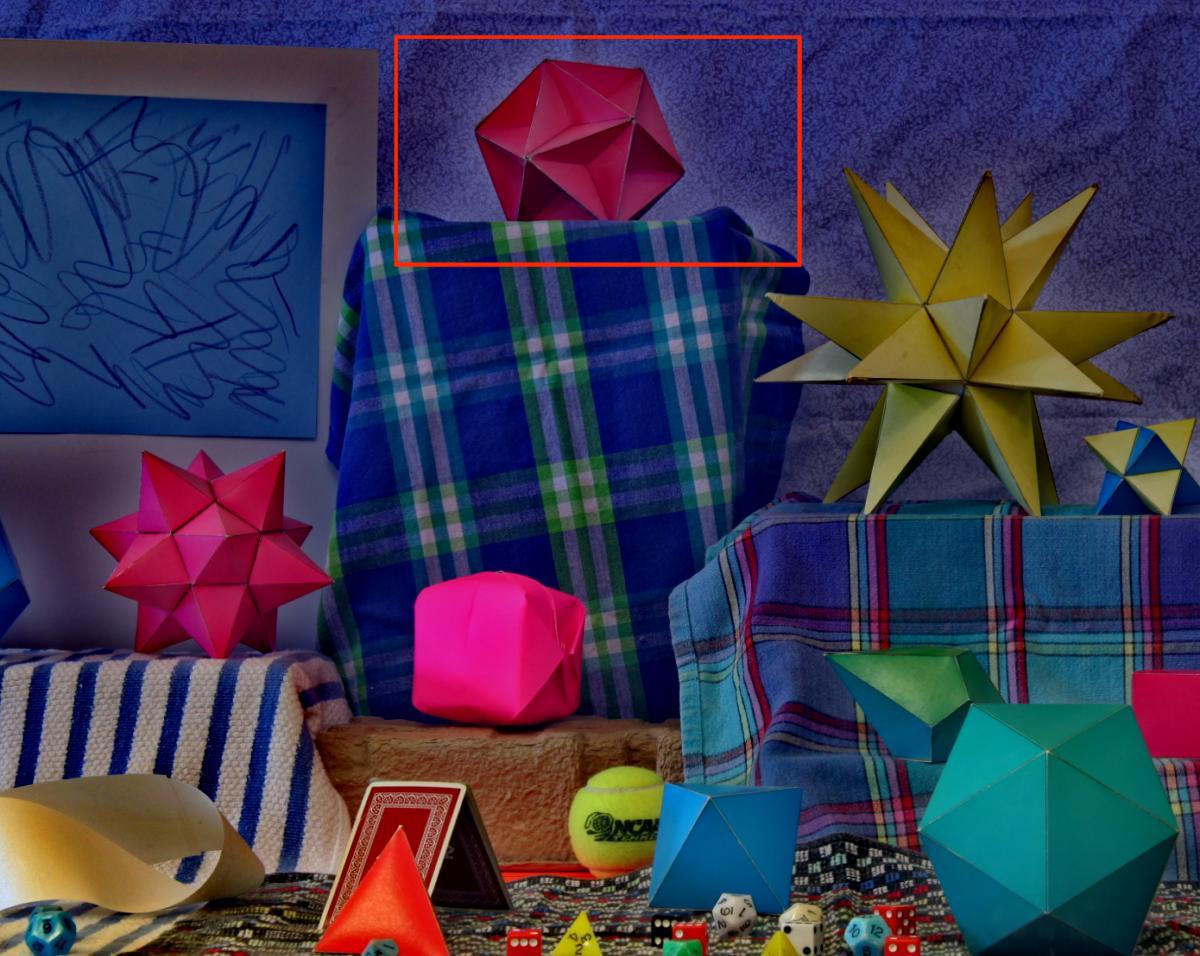}}\quad
\subfloat[Meng 's method~\cite{Meng}]{\includegraphics[width=.31\textwidth,height=3cm]{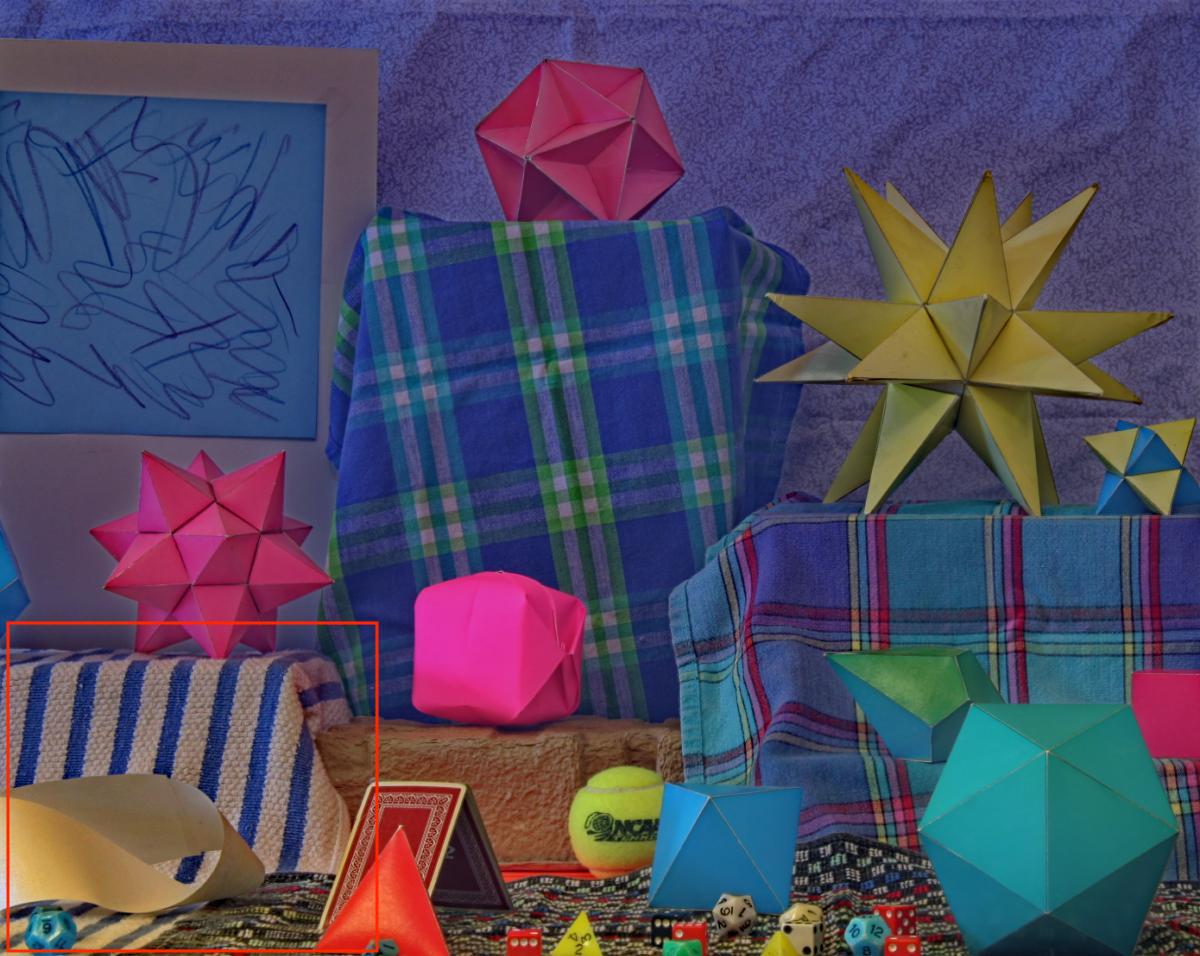}}\quad
\subfloat[Berman's method~\cite{Dana2016}]{\includegraphics[width=.31\textwidth,height=3cm]{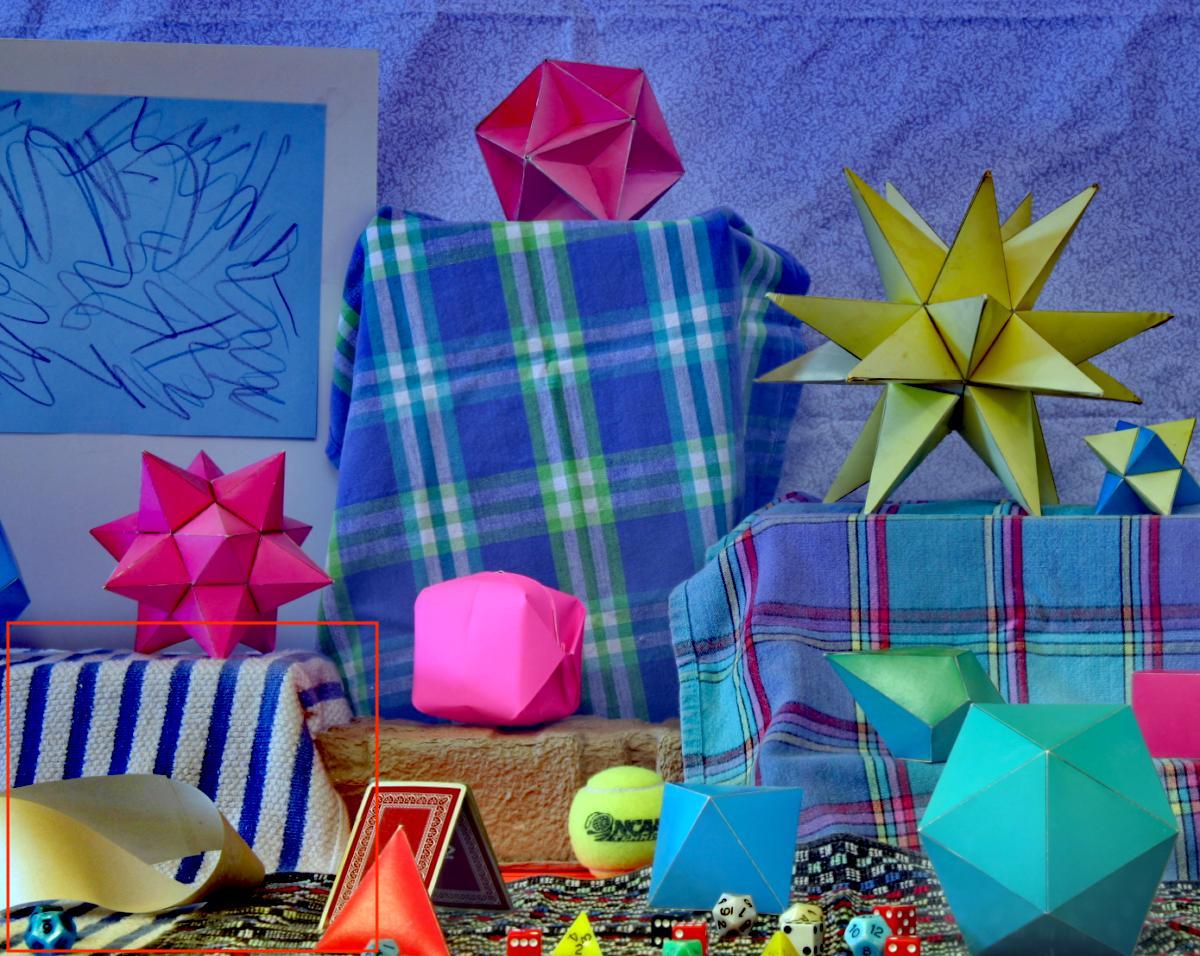}}\quad
\subfloat[Proposed method (MWTO)]{\includegraphics[width=.31\textwidth,height=3cm]{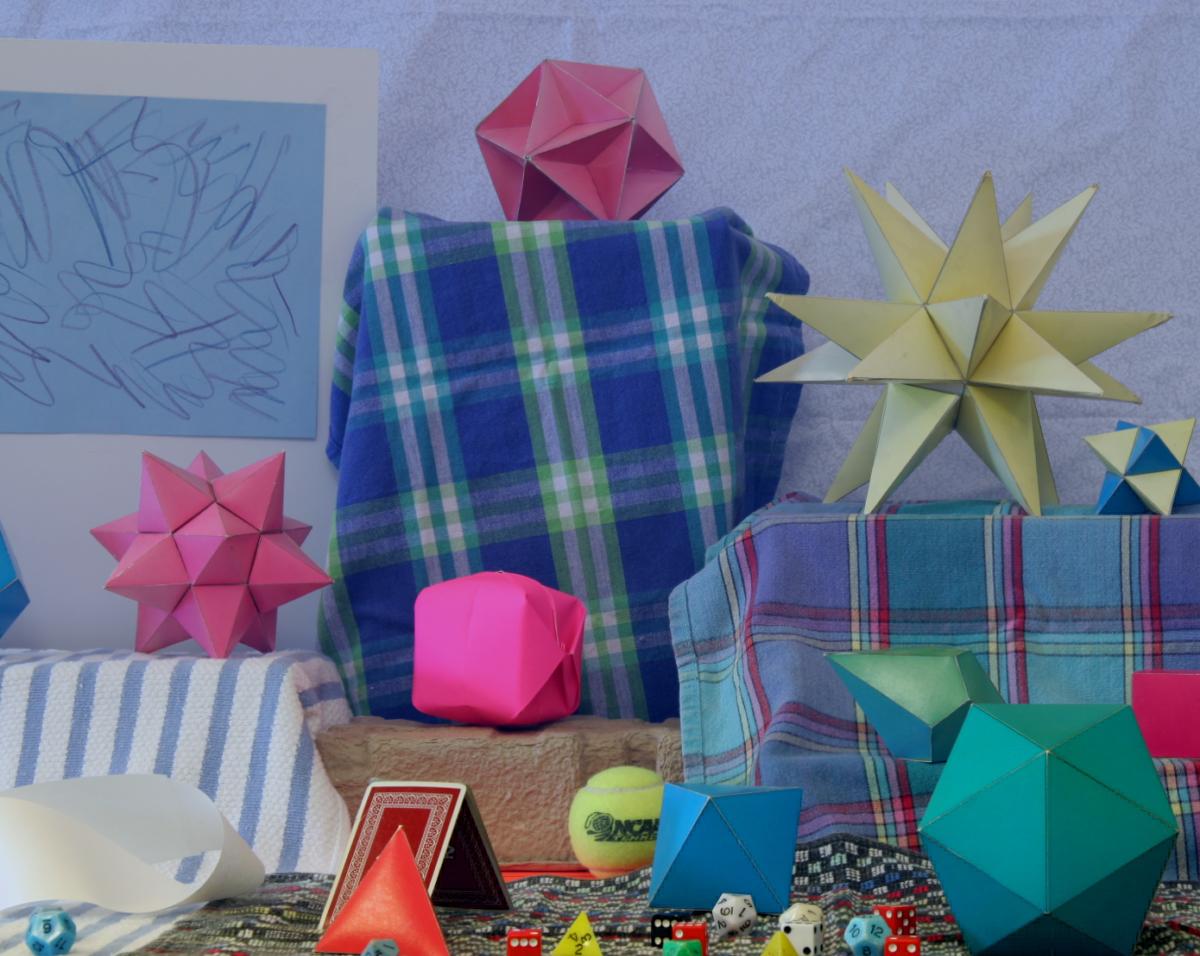}}\\
\caption{Dehazing results by different algorithms for the ``Desk" image.}
\label{fig:desk}
\end{center}
\end{figure*}

\begin{figure*}
\begin{center}
\subfloat[Hazy Input]{\includegraphics[width=.225\textwidth,height=4cm]{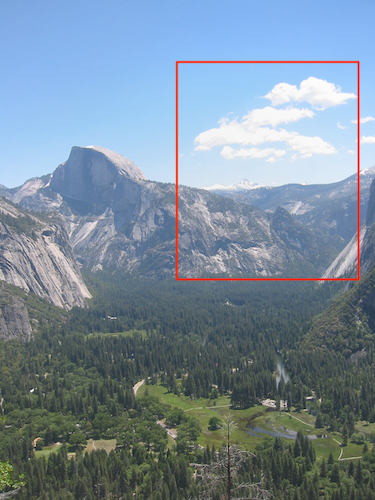}}\quad
\subfloat[Tarel's method~\cite{Tarel2009}]{\includegraphics[width=.225\textwidth,height=4cm]{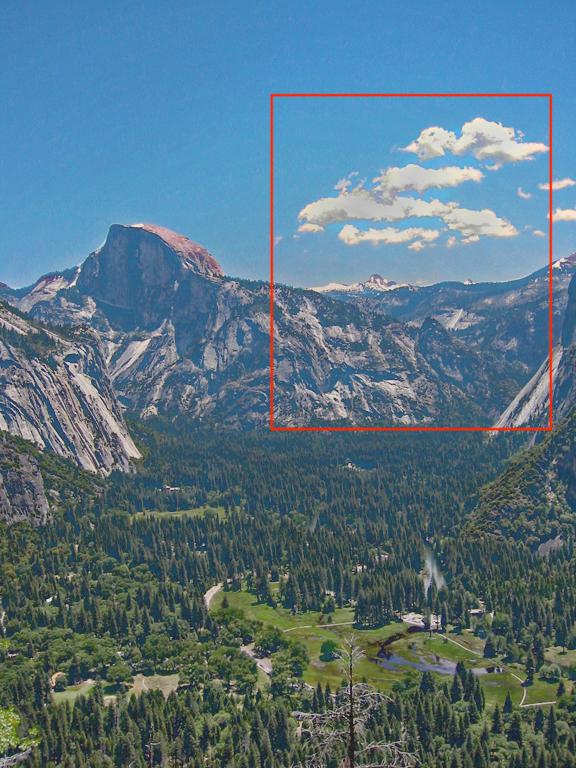}}\quad
\subfloat[He's method~\cite{Guided}]{\includegraphics[width=.225\textwidth,height=4cm]{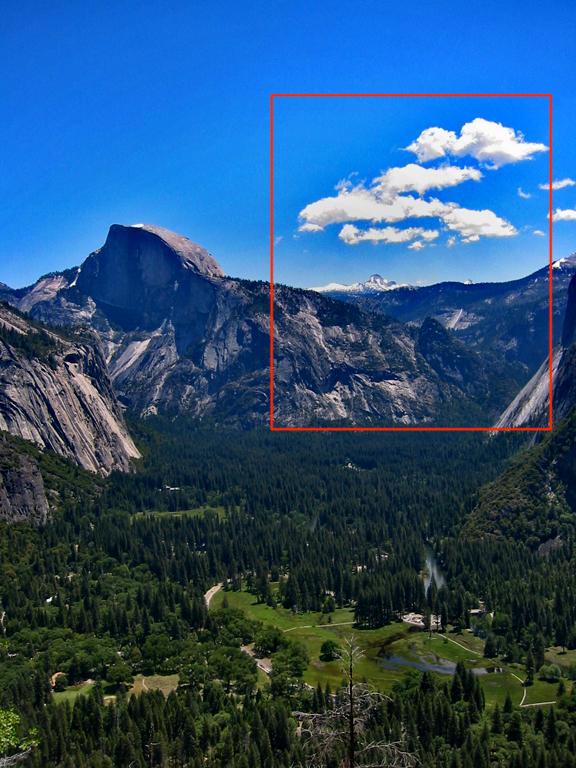}}\quad
\subfloat[Wang's method~\cite{Wang2014}]{\includegraphics[width=.225\textwidth,height=4cm]{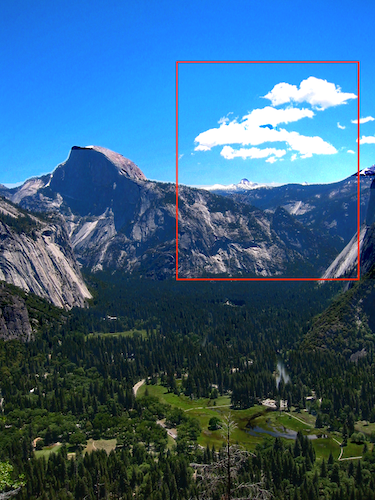}}\\
\subfloat[Meng's method~\cite{Meng}]{\includegraphics[width=.225\textwidth,height=4cm]{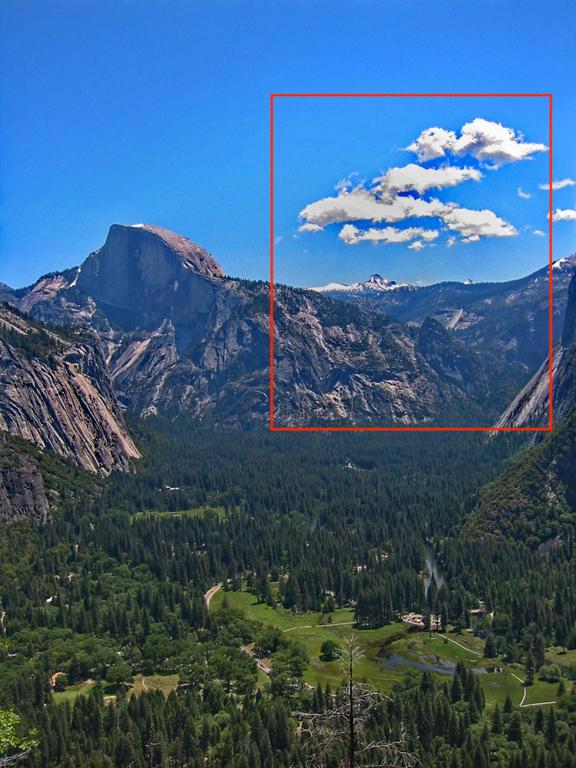}}\quad
\subfloat[Zhu's method~\cite{Zhu2015}]{\includegraphics[width=.225\textwidth,height=4cm]{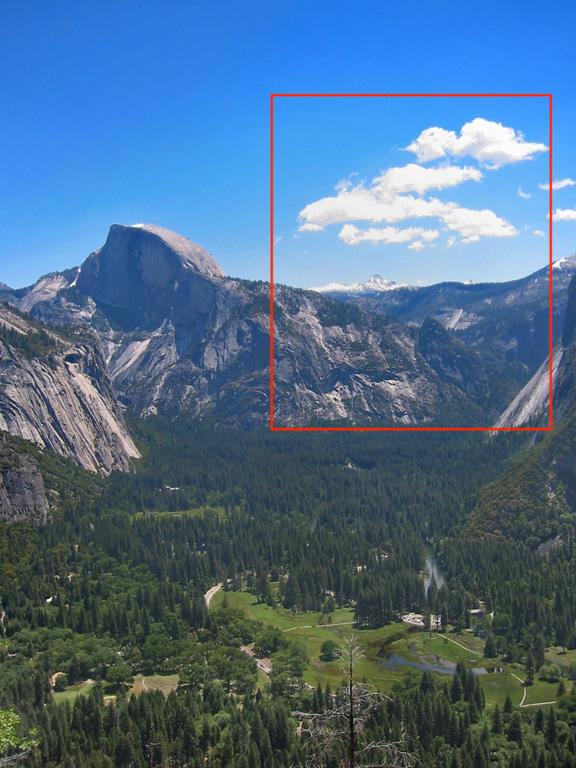}}\quad
\subfloat[Berman's~\cite{Dana2016}]{\includegraphics[width=.225\textwidth,height=4cm]{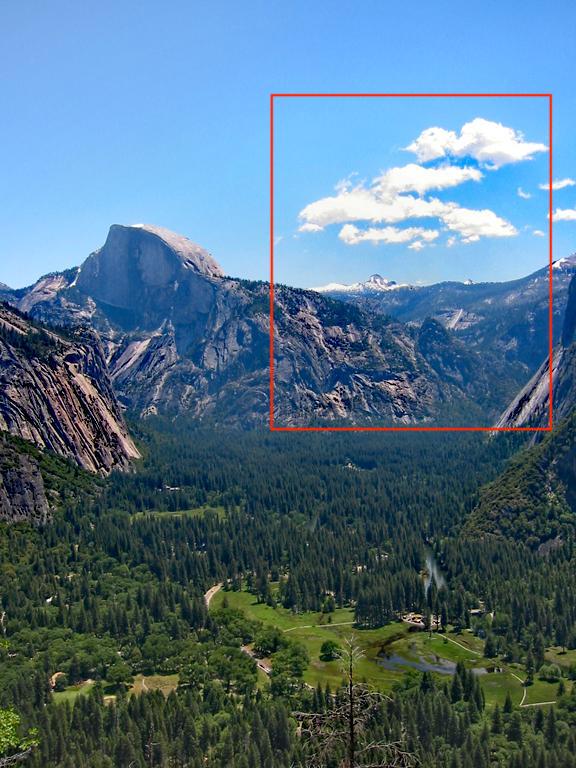}}\quad
\subfloat[Proposed (MWTO)]{\includegraphics[width=.225\textwidth,height=4cm]{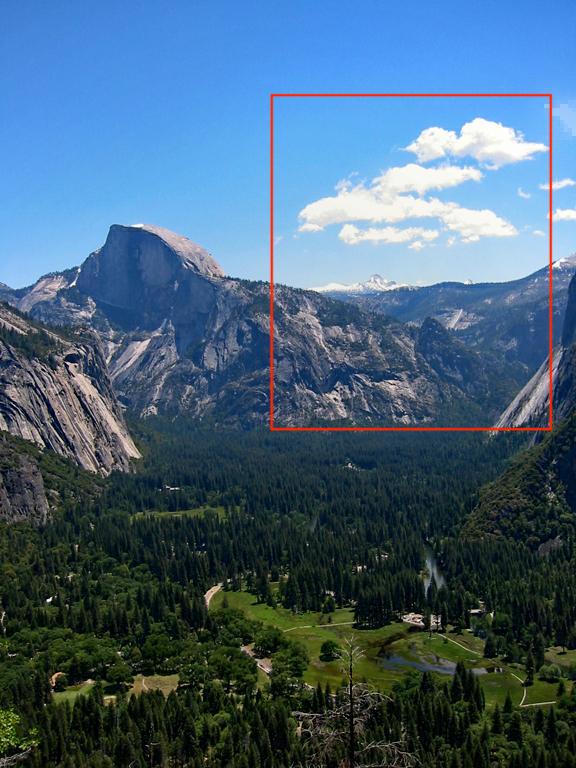}}\\
\caption{Dehazing results by different algorithms for the ``Hill" image.}
\label{fig:hill}
\end{center}
\end{figure*}

\begin{figure*}
\begin{center}
\subfloat[Hazy Input]{\includegraphics[width=.225\textwidth,height=3cm]{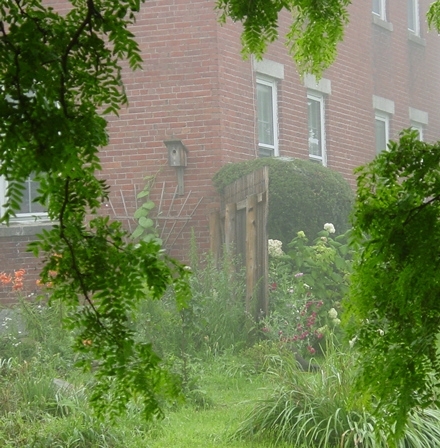}}\quad
\subfloat[Zhu's method~\cite{Zhu2015}]{\includegraphics[width=.225\textwidth,height=3cm]{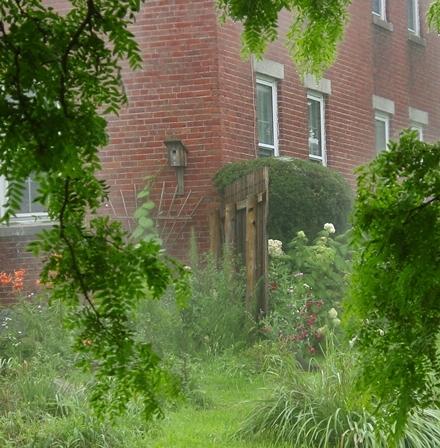}}\quad
\subfloat[He's method~\cite{Guided}]{\includegraphics[width=.225\textwidth,height=3cm]{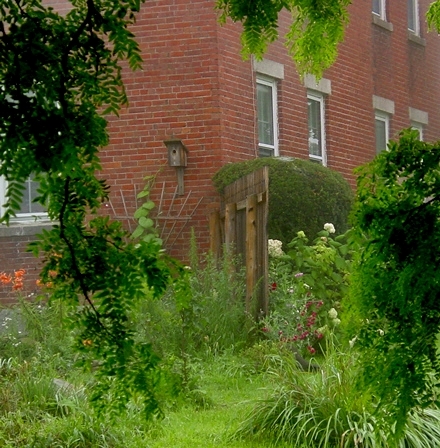}}\quad
\subfloat[Fattal's method~\cite{Fattal2008}]{\includegraphics[width=.225\textwidth,height=3cm]{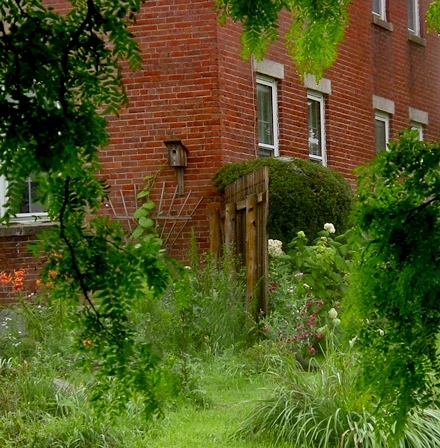}}\\
\subfloat[Nishion's~\cite{Bayesian-defog}]{\includegraphics[width=.225\textwidth,height=3cm]{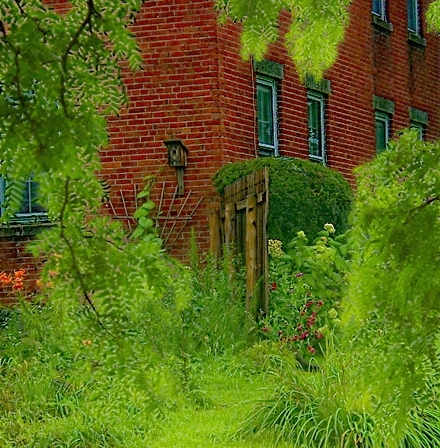}}\quad
\subfloat[Meng's method~\cite{Meng}]{\includegraphics[width=.225\textwidth,height=3cm]{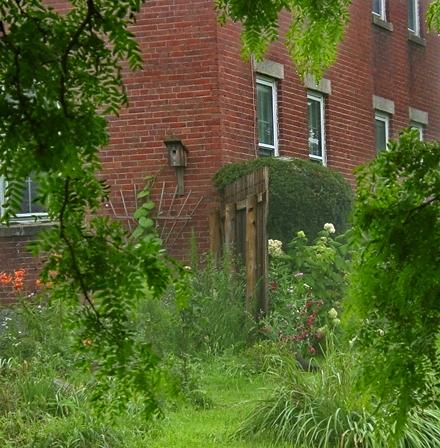}}\quad
\subfloat[Berman's~\cite{Dana2016}]{\includegraphics[width=.225\textwidth,height=3cm]{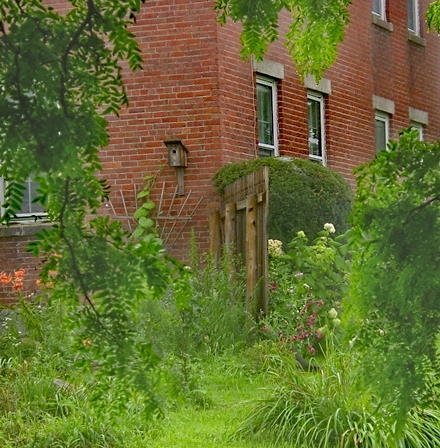}}\quad
\subfloat[Proposed (MWTO)]{\includegraphics[width=.225\textwidth,height=3cm]{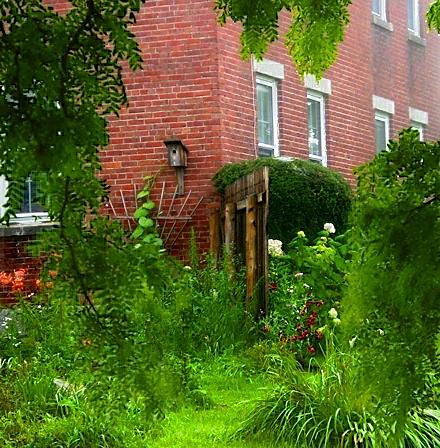}}\quad
\caption{Dehazing results by different algorithms for the ``House" image.}
\label{fig:house}
\end{center}
\end{figure*}

\begin{figure*}
\begin{center}
\subfloat[Hazy Input]{\includegraphics[width=.225\textwidth,height=3cm]{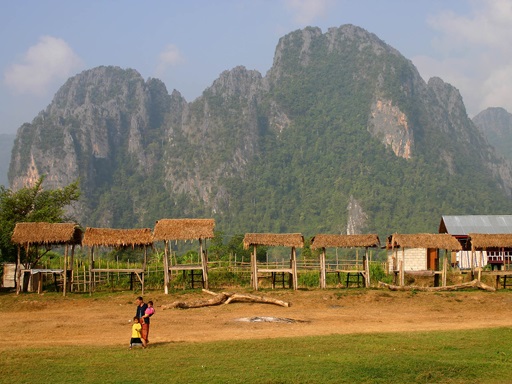}}\quad
\subfloat[Tarel's method~\cite{Tarel2009}]{\includegraphics[width=.225\textwidth,height=3cm]{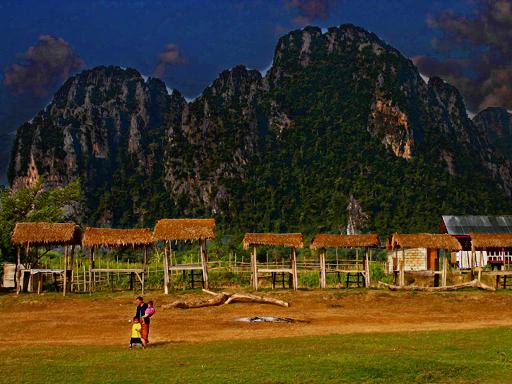}}\quad
\subfloat[He's method~\cite{Guided}]{\includegraphics[width=.225\textwidth,height=3cm]{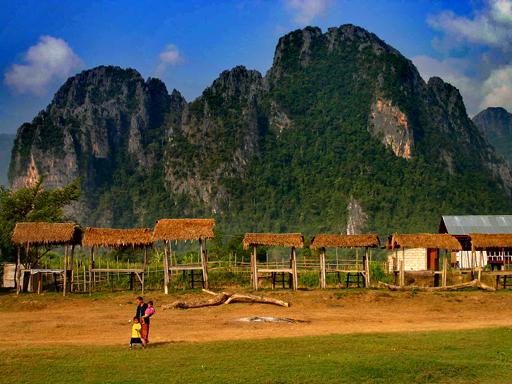}}\quad
\subfloat[Fattal's method~\cite{Fattal2008}]{\includegraphics[width=.225\textwidth,height=3cm]{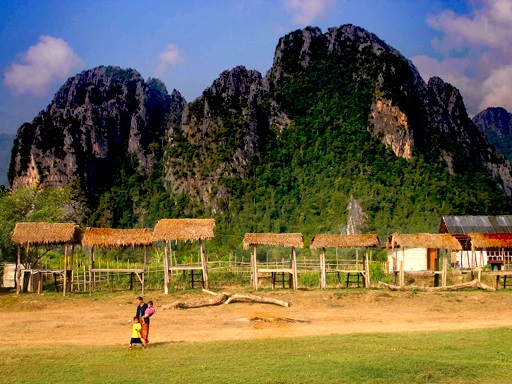}}\\
\subfloat[Meng's method~\cite{Meng}]{\includegraphics[width=.225\textwidth,height=3cm]{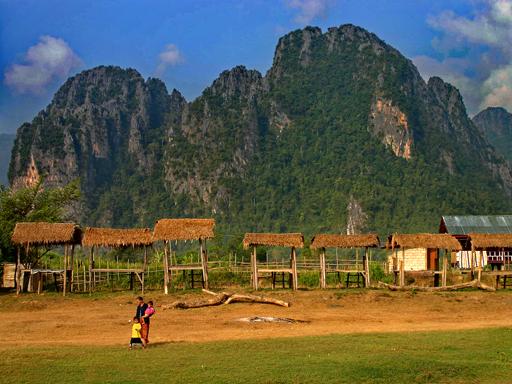}}\quad
\subfloat[Zhu's method~\cite{Zhu2015}]{\includegraphics[width=.225\textwidth,height=3cm]{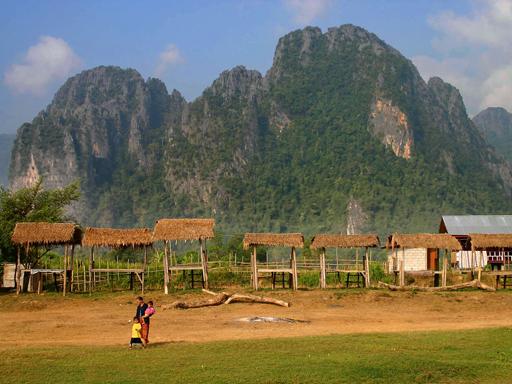}}\quad
\subfloat[Berman's~\cite{Dana2016}]{\includegraphics[width=.225\textwidth,height=3cm]{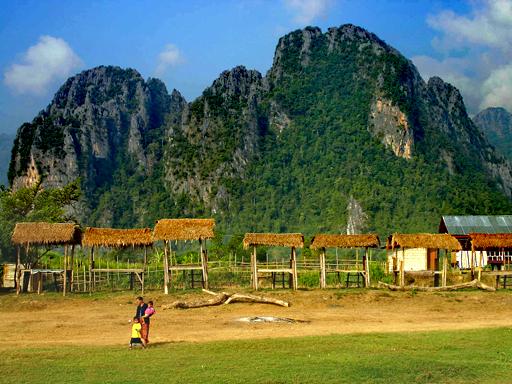}}\quad
\subfloat[Proposed (MWTO)]{\includegraphics[width=.225\textwidth,height=3cm]{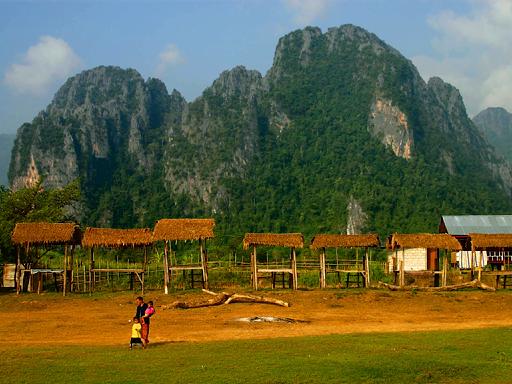}}\\
\caption{Dehazing results by different algorithms for the ``Mountain" image.}
\label{fig:Mountain}
\end{center}
\end{figure*}

\begin{figure*}[h]
\begin{center}
\subfloat[Hazy Input]{\includegraphics[width=.31\textwidth,height=3cm]{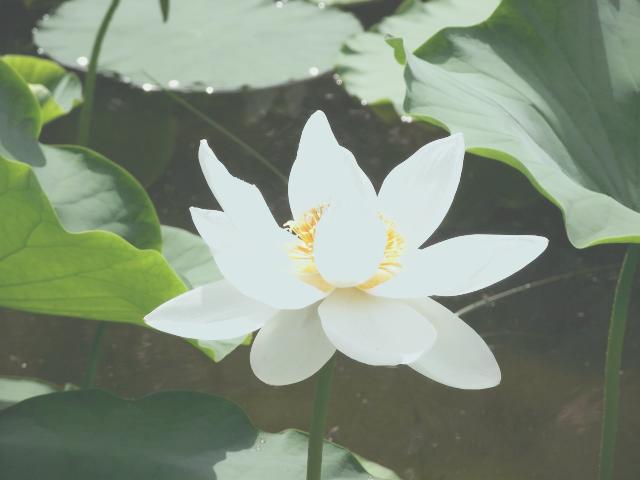}}\quad
\subfloat[He's method~\cite{Guided}]{\includegraphics[width=.31\textwidth,height=3cm]{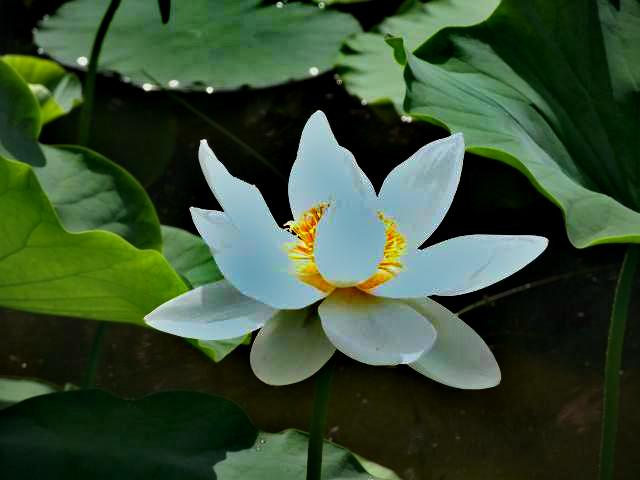}}\quad
\subfloat[Meng's method~\cite{Meng}]{\includegraphics[width=.31\textwidth,height=3cm]{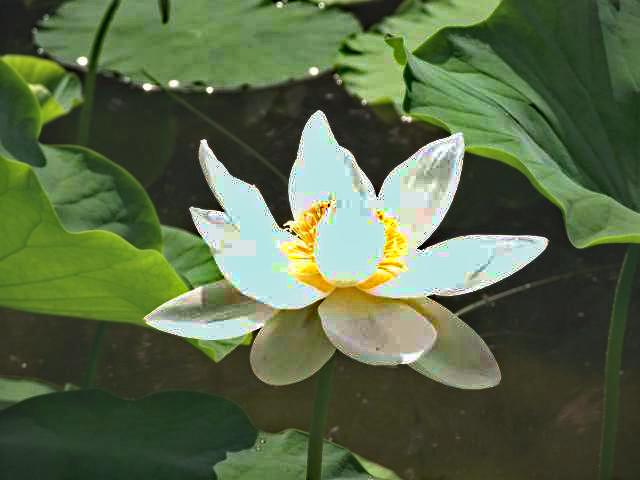}}\\
\subfloat[Zhu's method~\cite{Zhu2015}]{\includegraphics[width=.31\textwidth,height=3cm]{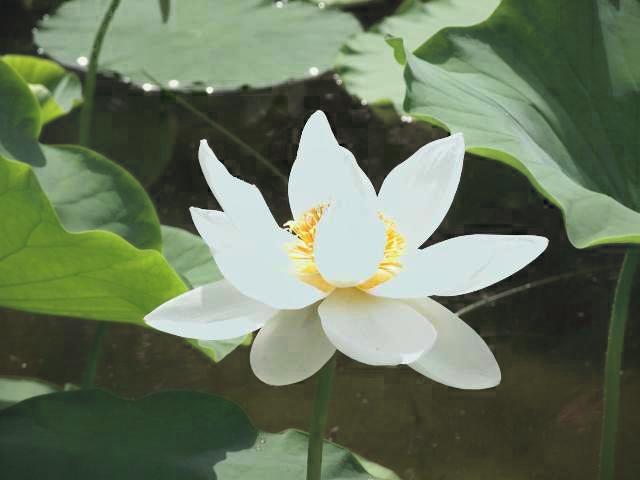}}\quad
\subfloat[Berman's method~\cite{Dana2016}]{\includegraphics[width=.31\textwidth,height=3cm]{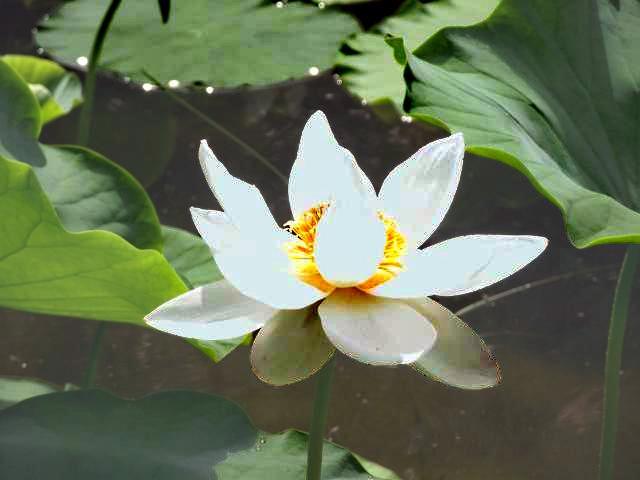}}\quad
\subfloat[Proposed (MWTO)]{\includegraphics[width=.31\textwidth,height=3cm]{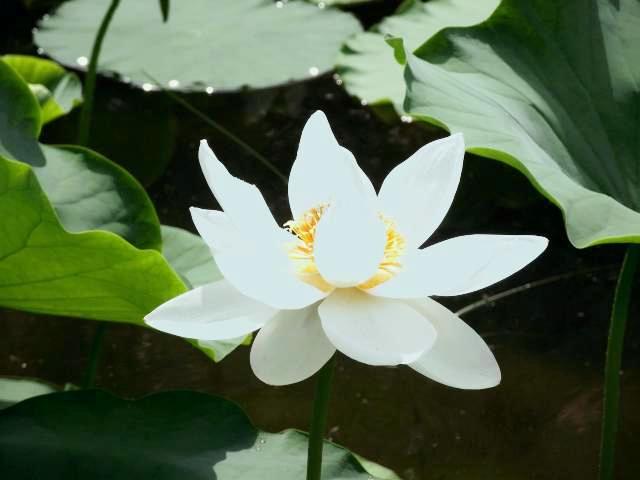}}\\
\caption{Dehazing results by different algorithms for the ``Lily" image.}
\label{fig:lily}
\end{center}
\end{figure*}

\begin{figure*}[h]
\begin{center}
\subfloat[Hazy Input]{\includegraphics[width=.31\textwidth,height=3cm]{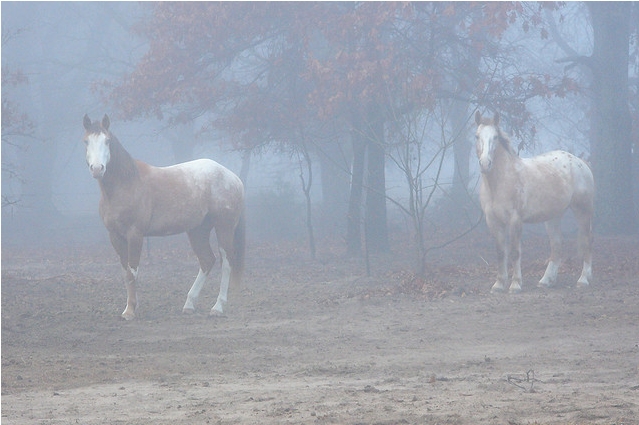}}\quad
\subfloat[Kim's method~\cite{Kim2013}]{\includegraphics[width=.31\textwidth,height=3cm]{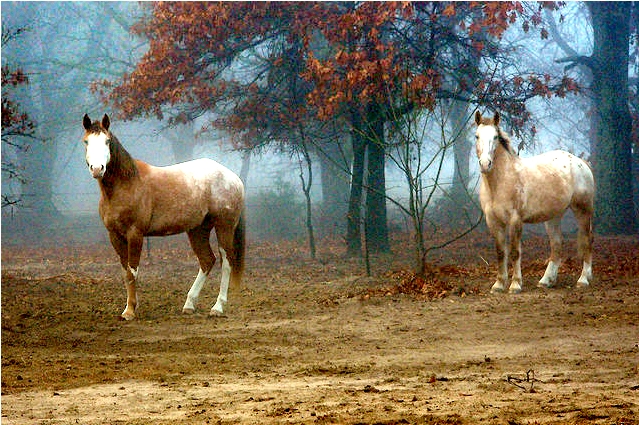}}\quad
\subfloat[Proposed (MWTO)]{\includegraphics[width=.31\textwidth,height=3cm]{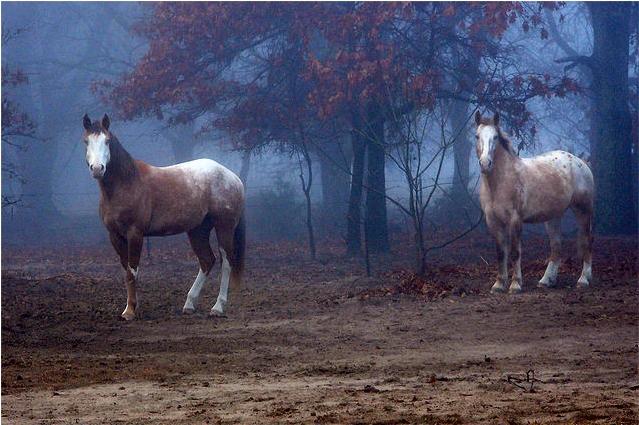}}\\
\caption{Dehazing results for the ``Horse" image.}
\label{fig:horse}
\end{center}
\end{figure*}

\begin{figure*}[h]
\begin{center}
\subfloat[Hazy Input]{\includegraphics[width=.225\textwidth,height=2.5cm]{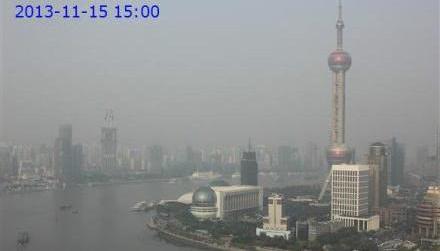}}\quad
\subfloat[Haze-free image]{\includegraphics[width=.225\textwidth,height=2.5cm]{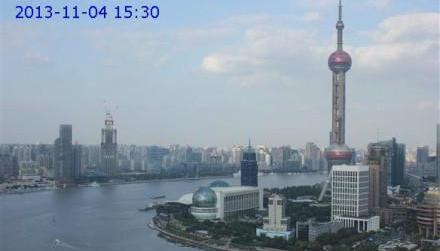}}\quad
\subfloat[Kim's method~\cite{Kim2013}]{\includegraphics[width=.225\textwidth,height=2.5cm]{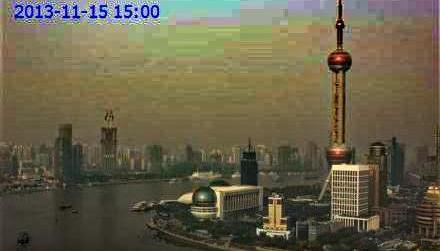}}\quad
\subfloat[He's method~\cite{Guided}]{\includegraphics[width=.225\textwidth,height=2.5cm]{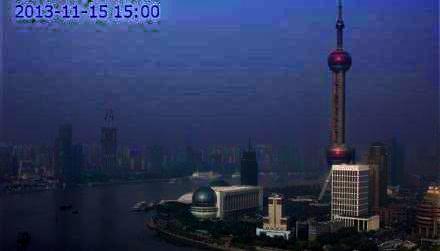}}\\
\subfloat[Meng's method~\cite{Meng}]{\includegraphics[width=.225\textwidth,height=2.5cm]{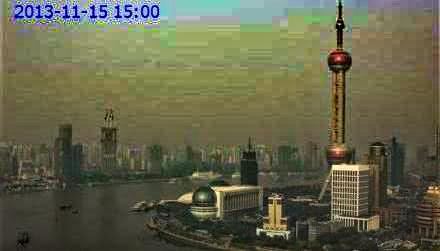}}\quad
\subfloat[Zhu's method~\cite{Zhu2015}]{\includegraphics[width=.225\textwidth,height=2.5cm]{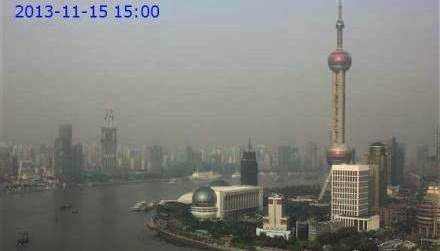}}\quad
\subfloat[Berman's~\cite{Dana2016}]{\includegraphics[width=.225\textwidth,height=2.5cm]{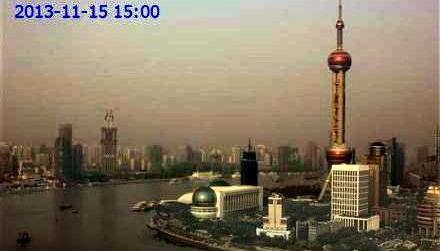}}\quad
\subfloat[Proposed (MWTO)]{\includegraphics[width=.225\textwidth,height=2.5cm]{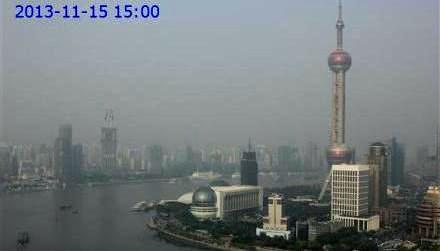}}\\
\caption{Dehazing results by different algorithms for the ``River" image.}
\label{fig:River}
\end{center}
\end{figure*}

\begin{figure*}
\begin{center}
\subfloat[Hazy Input]{\includegraphics[width=.225\textwidth,height=2.5cm]{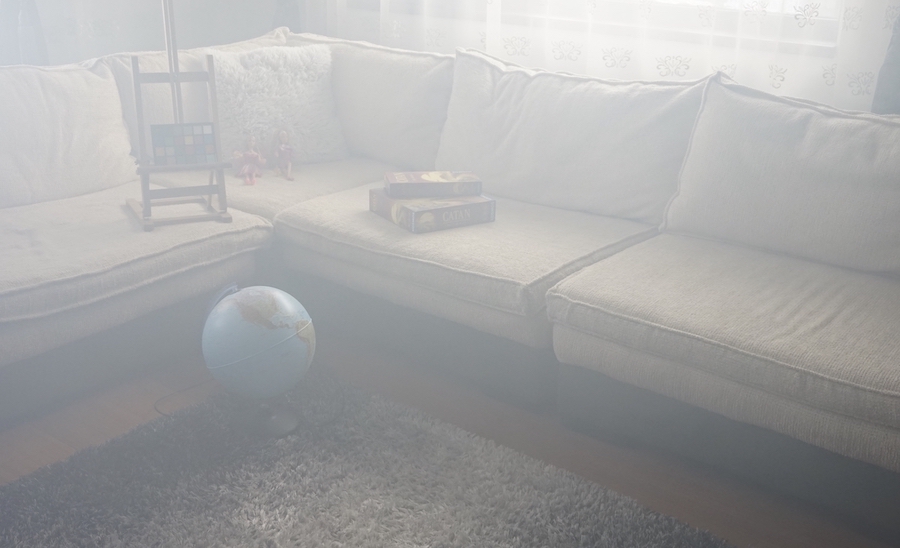}}\quad
\subfloat[Haze-free image]{\includegraphics[width=.225\textwidth,height=2.5cm]{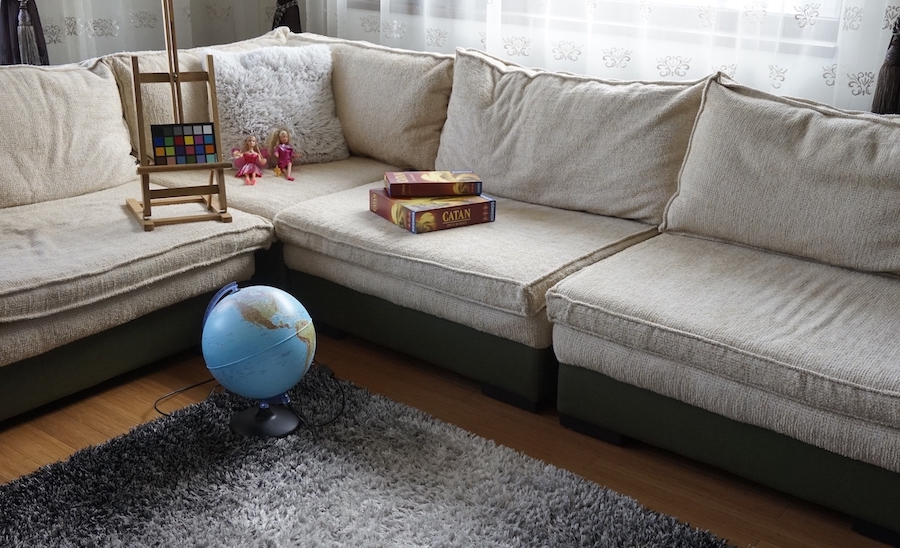}}\quad
\subfloat[He's method~\cite{Guided}]{\includegraphics[width=.225\textwidth,height=2.5cm]{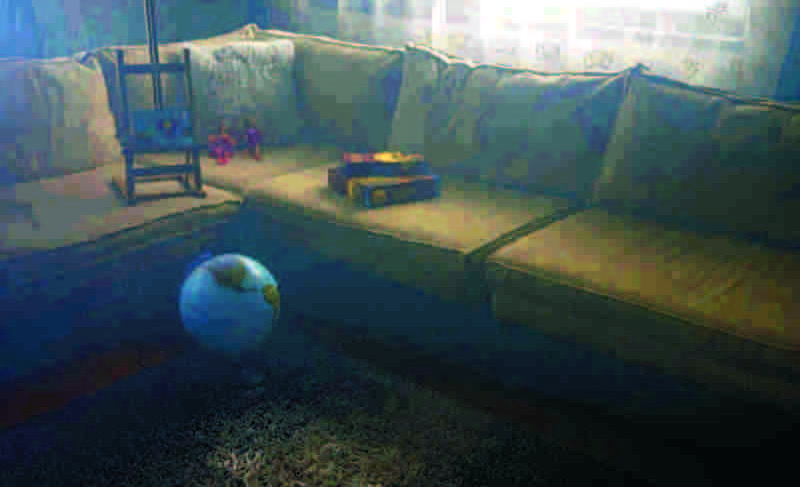}}\quad
\subfloat[Meng's method~\cite{Meng}]{\includegraphics[width=.225\textwidth,height=2.5cm]{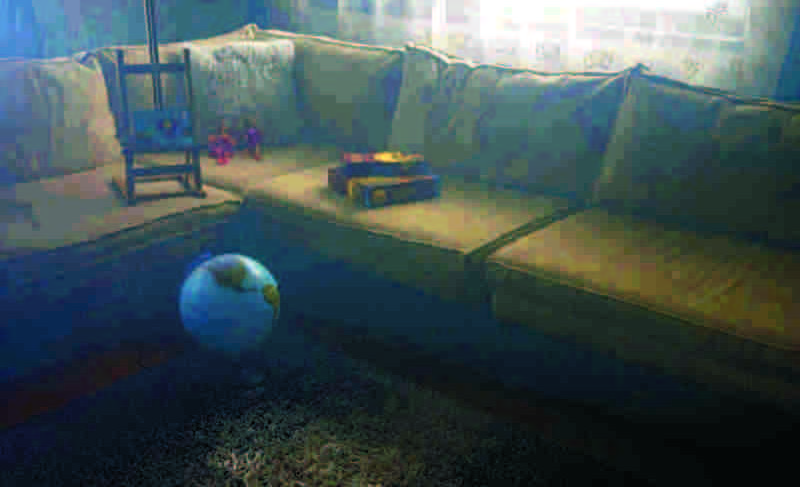}}\\
\subfloat[Ancuti method~\cite{airlight}]{\includegraphics[width=.225\textwidth,height=2.5cm]{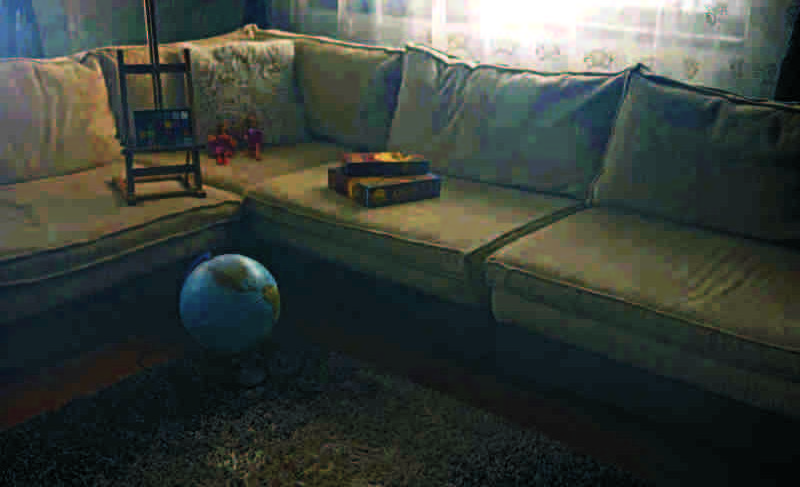}}\quad
\subfloat[Fattal's method~\cite{Fattal2008}]{\includegraphics[width=.225\textwidth,height=2.5cm]{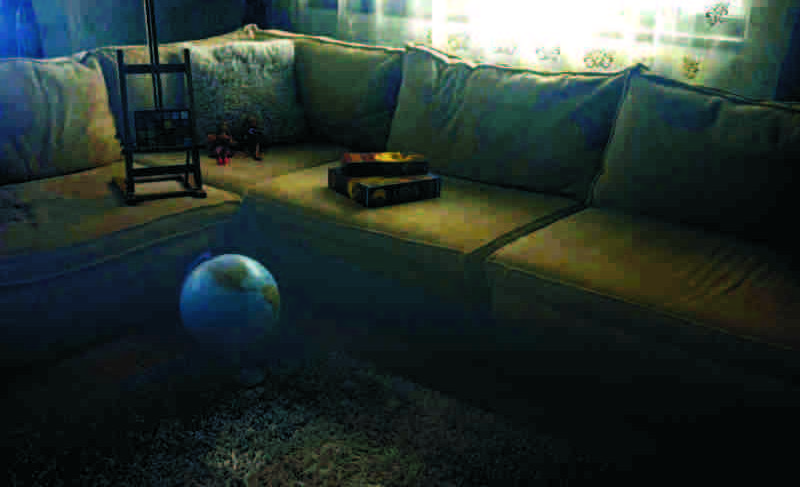}}\quad
\subfloat[Berman method~\cite{Dana2016}]{\includegraphics[width=.225\textwidth,height=2.5cm]{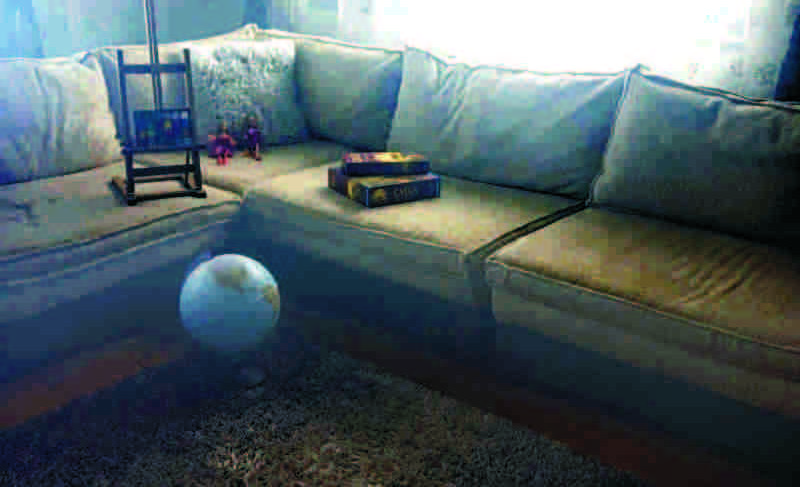}}\quad
\subfloat[Proposed (MWTO)]{\includegraphics[width=.225\textwidth,height=2.5cm]{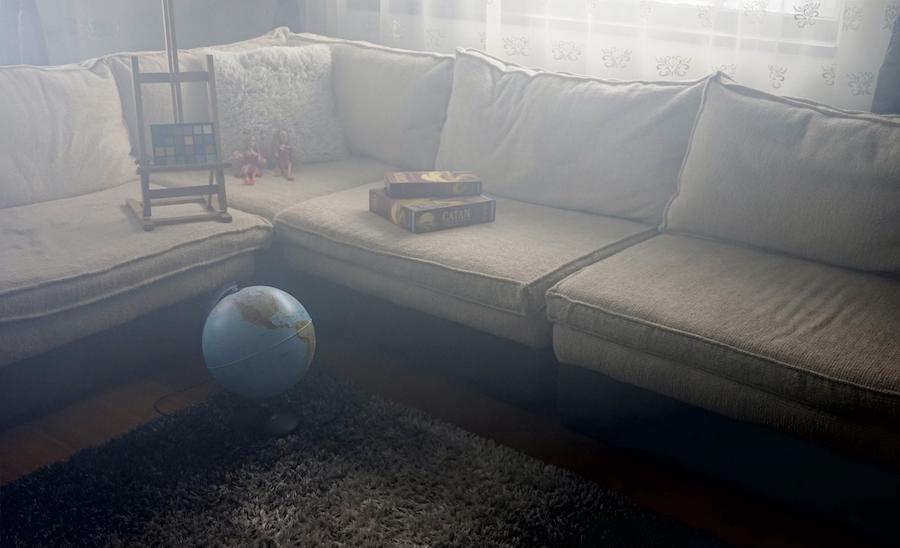}}\\
\caption{Dehazing results by different algorithms for the ``Sofa" image.}
\label{fig:sofa}
\end{center}
\end{figure*}

\begin{figure*}
\begin{center}
\subfloat[Hazy Input]{\includegraphics[width=.225\textwidth,height=2.5cm]{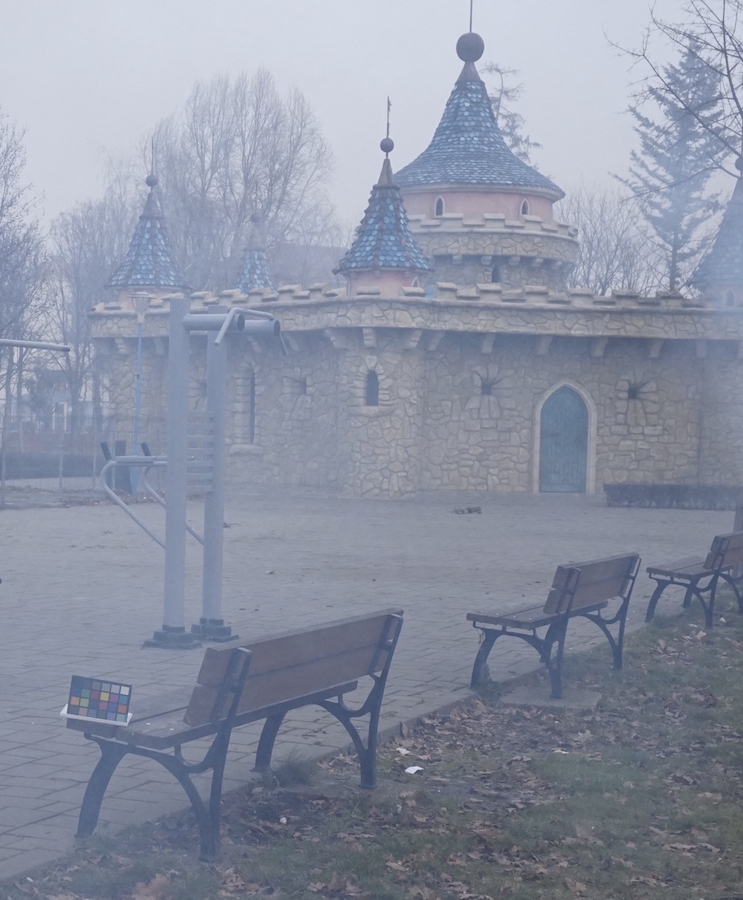}}\quad
\subfloat[Haze-free image]{\includegraphics[width=.225\textwidth,height=2.5cm]{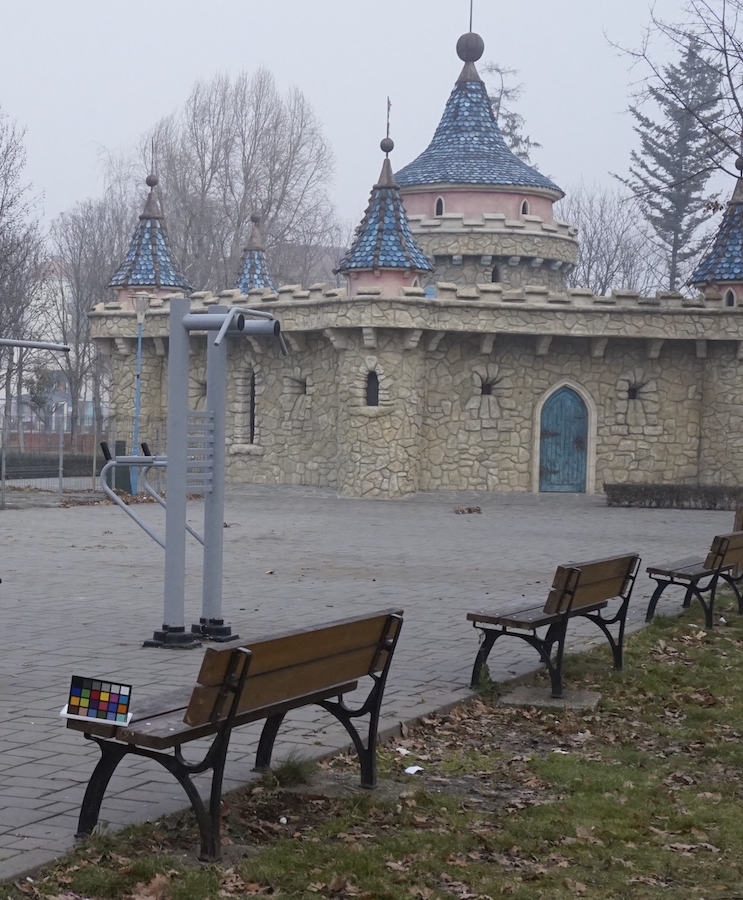}}\quad
\subfloat[He's method~\cite{Guided}]{\includegraphics[width=.225\textwidth,height=2.5cm]{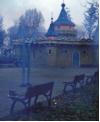}}\quad
\subfloat[Meng's method~\cite{Meng}]{\includegraphics[width=.225\textwidth,height=2.5cm]{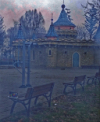}}\\
\subfloat[Ancuti method~\cite{airlight}]{\includegraphics[width=.225\textwidth,height=2.5cm]{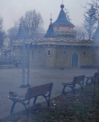}}\quad
\subfloat[Fattal's method~\cite{Fattal2008}]{\includegraphics[width=.225\textwidth,height=2.5cm]{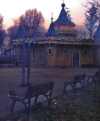}}\quad
\subfloat[Berman method~\cite{Dana2016}]{\includegraphics[width=.225\textwidth,height=2.5cm]{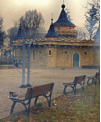}}\quad
\subfloat[Proposed (MWTO)]{\includegraphics[width=.225\textwidth,height=2.5cm]{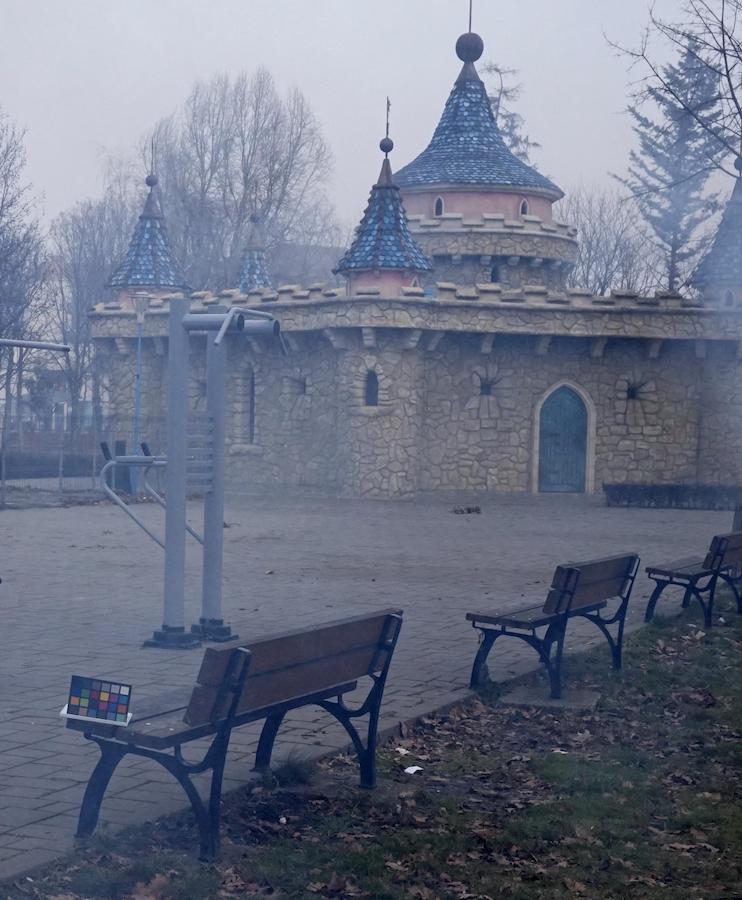}}\\
\caption{Dehazing results by different algorithms for the ``Castle" image.}
\label{fig:castle}
\end{center}
\end{figure*}

In the classic problem setting of image dehazing, only the hazy image is provided and the ``ground truth" image is hard to obtain. Fig.~\ref{fig:Canyon},~\ref{fig:desk},~\ref{fig:hill},~\ref{fig:house},~\ref{fig:Mountain}, and~\ref{fig:horse} compare the experimental result (dehazed image) of the proposed method (MWTO) with the algorithms presented in~\cite{Tarel2009, Guided, Meng, Kim2013, Zhu2015, Dana2016}. Additionally, we present dehazing result of Fig.~\ref{fig:lily}, where simulated haze is applied to the original image;~Fig.~~\ref{fig:sofa},~\ref{fig:castle} from two recently published datasets~\cite{ihaze,ohaze} and~Fig.~\ref{fig:River}, where the ``ground-truth" photograph taken on a haze-free day is available. 

It can be observed that our proposed method usually produces a more faithful and balanced contrast and better color reconstruction over the whole image. In the ``Canyon" image, there are fringe artifacts in the result of Tarel's algorithm~\cite{Tarel2009}. He's method~\cite{Guided} results in the most oversaturated effect, because it tends to underestimate the transmission function. Meng's method~\cite{Meng} attempts to improve He's method~\cite{Guided} by including a boundary constraint for restoring bright color and limiting over-saturation. However, this constraint does not work well in areas such as sky and cloud. Berman's method~\cite{Dana2016} seems better than the results of~\cite{Guided} and~\cite{Meng}, though suffers from a similar over-saturation effect. 


This over-saturation problem is common and also presented in the ``Desk" image and the ``Sofa" image. He's method~\cite{Guided}, Meng's method~\cite{Meng}, and Berman's method~\cite{Dana2016} show over-correction of color contrast, especially on the objects at the lower left corner of the ``Desk" image. Halos around the pink hexagonal object in the middle is salient, which may be caused by defects of the DCP technique~\cite{Guided} and the hazy-line technique~\cite{Dana2016} since pixel-wise based estimation does not consider the continuity of adjacent pixels. Zhu's method~\cite{Zhu2015} does not have the problem of over-saturation and halos, however, the dehazing effect is minimally perceivable. In contrast, the result of MWTO is satisfying for the ``Desk" in terms of hue and detail reconstruction. In the ``Sofa" image, He's method~\cite{Guided}, Meng's method~\cite{Meng}, Ancuti's method~\cite{airlight}, and Fattal's method~\cite{Fattal2008} all suffer from the overcorrected, darkened effect. Only Berman's method~\cite{Dana2016} produces a result close to the ground truth, but shows a surrealistic outlook. Our proposed method does not completely remove the haze, but still preserve the correct saturation and details.

In the ``Hill" image, Tarel's algorithm~\cite{Tarel2009} again introduces heavy color distortion and undesired artifacts. He's method~\cite{Guided} and Wang's method~\cite{Wang2014} cause over-saturation and loss of detailed information (see the over-exposed and whitening effect of cloud in Wang's method~\cite{Wang2014}). Berman's method\cite{Dana2016} produces a nice color, however, the over shapening around the left hill and between gaps of cloud makes it visually unreal. Zhu's method~\cite{Zhu2015} is unable to thoroughly remove the haze, especially on the central forest. In contrast, our result is pleasing in terms of visual effect and local details. In the ``house" image, many methods generate halos or over-saturation around the left tree branch~\cite{Fattal2008,Guided,Kim2013, Meng, Zhu2015, Dana2016}. While our proposed method offers good visual effect with color enhanced and information preserved. Fig.~\ref{fig:Mountain} shows that DCP based techniques can severely darken the image when there is a heavy haze~\cite{Tarel2009, Guided, Fattal2008}. In this situation, Zhu's method~\cite{Zhu2015} again fails to sufficiently remove the haze and Meng's method~\cite{Meng} and Berman's method~\cite{Dana2016} cause over-saturation in the sky area. In comparison with those methods, our proposed method presents more natural color and a satisfying contrast. 

In addition to the natural hazed images, we further investigate simulated haze, which can be very useful in understanding haze formation and virtual scene manipulation. Fig.~\ref{fig:lily} addresses the advantage of having an optical image model when the haze is simulated. We notice that Zhu's method~\cite{Zhu2015} can not completely remove the haze. Meng's method~\cite{Meng} brings about artifacts and a slight color distortion. He's method~\cite{Guided}, Meng's method~\cite{Meng}, and Berman's method~\cite{Dana2016} all cause shadow on the lower petals, which can be avoided if transmission map is estimated with a physically sound model. Our method can effectively solve the inverse problem of haze formation and produce a close estimate to the haze-free image. 



The color distortion problem can be sometimes severe, especially in the contrast enhancement based methods because bright colors are more visually perceivable. For instance in Fig.~\ref{fig:horse}, the hue of ground and horses is completely changed by Kim's method~\cite{Kim2013}. In Figure~\ref{fig:castle}, Fattal's method~\cite{Fattal2008} and Berman's method~\cite{Dana2016} significantly change the color as well. In Fig.~\ref{fig:River}, given the haze-free image in comparison, Kim's method~\cite{Kim2013}, He's method~\cite{Guided}, and Berman's method~\cite{Dana2016} all produce a rather yellowish, distorted result. Zhu's method~\cite{Zhu2015} tackles the color distortion problem by using color attenuation prior. Recently, Lian's method~\cite{Lian-18} considers the detail mapping separately from the global intensity mapping of haze. These two methods do not suffer from the color problem. But in terms of the perceived naturalness, our model is still among the best of the compared methods (see Fig.~\ref{fig:River}).

\subsection{Objective evaluation}
Unlike subjective evaluation which is manually conducted and can introduce bias, we also quantitatively evaluate the quality of dehazed image. One such method is to set up a reference image by optimizing all the parameters of the dehazing algorithm, and calculate the distance between the dehazed image and the reference image. We consider MSE as the distance measure (calculated as in equation~\ref{eq:mse}, where $M$ and $N$ are image dimensions, $f(i,j)$ is the reference image and $f^\prime(i,j)$ the dehazed image). Other distance measures, such as the peak signal-to-noise ratio (PSNR) and structural similarity (SSIM) are also mentioned in the literature~\cite{wangw17}. It is worth mentioning that these metrics only serve as references and are often not well-aligned with the visual effects. Another well-known quantitative evaluation is Hautiere's method~\cite{Tarel2008}, which defines three measures for visibility: the visible edge ratio $e$, percentage of saturated pixels in all color channels $\sigma$, and the visible edge normalized gradient $\bar{r}$ as in~(\ref{eq:esr}).\\ 
\begin{equation}
\label{eq:mse}
\text{MSE}=\frac{1}{M\times N} \sum_{i=0}^{M-1}\sum_{j=0}^{N-1}[f(i,j)-f^\prime(i,j)].
\end{equation}
The numbers of visible edges in the original image and dehazed image are denoted by $n_o$ and $n_r$ respectively. $P_i$ denotes pixels on the visible edges of the dehazed image and $r_i$ the corresponding gradient. $n_s$ is the number of saturated (balck or white) pixels. 
\begin{align} \label{eq:esr}
e=& \frac{n_r -n_o}{n_o},\nonumber\\
\sigma=& \frac{n_s}{\dim_x\times\dim_y},\nonumber\\
\bar{r}=& \exp[\frac{1}{n_r} \sum_{P_i \in \wp_r} \log r_i].
\end{align}

Table~\ref{t:hau} reports the average Hautiere's visibility descriptors as well as the MSE calculated from the ``Lily" image mentioned in Section~\ref{sec:results}. Our method outperforms all the others including He's method~\cite{Guided}, Kim's method~\cite{Kim2013}, Meng's method~\cite{Meng}, Zhu's method~\cite{Zhu2015}, and Berman's method~\cite{Dana2016} in terms of MSE and achieves the second best results in terms of $e$, $\sigma$, and $\bar{r}$. This observation confirms the right balance and stable performance of MWTO. It is also important to notice that these results are obtained with a much faster computation speed.

We also report the average PSNR and SSIM obtained across the I-HAZE~\cite{ihaze} and O-HAZE~\cite{ohaze} datasets in Table~\ref{psnrssim}. We find out that our method works better on indoor scenes compared to outdoor scenes.

\begin{table}[h]
\centering
\caption{Visual quality descriptors of compared methods.} \label{t:hau}
\begin{tabular}{lp{1.85cm}p{1.85cm}p{1.85cm}}
\toprule
 & He's~\cite{Guided} & Kim's~\cite{Kim2013}& Meng's~\cite{Meng}\\ 
\midrule
$e$ & 1.2& 0.7 & 2  \\
$\sigma$ & 0.7& 0.8 & 0.06 \\
$\bar{r}$ & 1.4 & 1.3 &  \textbf{2}\\
MSE & 0.51 & 0.44 & 0.44 \\
\toprule
& Zhu's~\cite{Zhu2015} & Berman's~\cite{Dana2016} & Our method\\
\midrule
$e$ & 1.2 & \textbf{1.8} & \textbf{1.4} \\
$\sigma$ & \textbf{0} & 0.6 & \textbf{0.05} \\
$\bar{r}$  & 0.8 & 1.4 & \textbf{1.6}\\
MSE  & 0.33 & 0.34 & \textbf{0.31}\\
\bottomrule
\end{tabular}
\end{table}

\begin{table}[h]
\centering
\caption{Quantitative evaluation results across the entire I-HAZE and O-HAZE datasets.} \label{psnrssim}
\resizebox{\textwidth}{!}{
\begin{tabular}{lrrrrrr}
\toprule
Metrics &He's~\cite{Guided} & Meng's~\cite{Meng} & Fattal's~\cite{Fattal2008} & Ancuti's~\cite{airlight} & Berman's~\cite{Dana2016} & Our method\\ 
\midrule
PSNR-indoor & 15.285 & 14.574 & 12.421 & \textbf{16.632} & 15.942 & \textbf{16.619} \\
SSIM-indoor & 0.711 & 0.750 & 0.574 & \textbf{0.770} & \textbf{0.767} & 0.619 \\
\midrule
PSNR-outdoor & 16.586 & \textbf{17.443} & 15.630 & \textbf{16.855} & 16.610 & 15.347  \\
SSIM-outdoor & 0.735 & \textbf{0.753} & 0.707 & 0.747 & \textbf{0.750} & 0.392 \\
\bottomrule
\end{tabular}}
\end{table}

\begin{table}[h]
\centering
\caption{Computational time (second) of different algorithms for experimented images.} \label{t:compare_time1}
\begin{tabular}{lrrr}
\toprule
Algorithm & Canyon & Desk & Hill\\ 
 & \small{(707$\times$565)} & \small{(1200$\times$956)} & \small{(576$\times$768)}\\ 
\midrule
He's~\cite{Guided} & 14.5 & 24.4 & 17.2\\
Meng's~\cite{Meng} & 3.8 & 9.4 & 3.9\\
Zhu's~\cite{Zhu2015}  & 2.8& 4.1 & 3.8\\
Berman's~\cite{Dana2016} & 2.9 & 6.4 & 3.0\\
Our method &  \textbf{0.8} &  \textbf{1.6} &  \textbf{0.8}\\
\toprule
Algorithm & House & Lily & River \\ 
& \small{(440$\times$448)}& \small{(640$\times$480)}&\small{(440$\times$260)}\\ 
\midrule
He's~\cite{Guided} & 7.8 & 12.0 & 4.3 \\
Meng's~\cite{Meng} & 2.8 & 3.5 & 2.2\\
Zhu's~\cite{Zhu2015}  & 2.1 & 3.0 & 1.8 \\
Berman's~\cite{Dana2016} & 1.6 & 2.1& 1.1\\
Our method & \textbf{0.5} &  \textbf{0.7} & \textbf{0.4}\\
\bottomrule
\end{tabular}
\end{table}

\subsection{Computational complexity}
The fast processing speed is a key advantage of MWTO. Unlike most of the compared methods which use filters to accelerate the processing speed, our method leverages DHWT, which preserves the maximum information and significantly speed up the dehazing process. We conduct experiments on 10 images and observe \emph{consistent} results. Table~\ref{t:compare_time1} reports 6 out of them due to space limit. These algorithms are tested on the same processor using the same MATLAB configuration.
He's method~\cite{Guided}, though used a guided filter instead of the soft-matting technique in its predecessor~\cite{He2011}, does not seem favorable in terms of speed. Meng's method~\cite{Meng} and Zhu's method~\cite{Zhu2015} are faster, which may be the result of employing a linear model. Berman's method~\cite{Dana2016} further improves the processing speed in average. However, it is noticeable that our method with two-level DHWT is already faster than Berman's method~\cite{Dana2016} and only requires $28\%$ of its processing time in average. By applying a larger multi-level parameter, our method can be further accelerated with a minimum trade-off against quality. We have successfully deployed this algorithm for real-time processing of $352\times240$ videos with a higher computational power.

\begin{figure}[t]
\begin{center}
\includegraphics[width=0.9\textwidth]{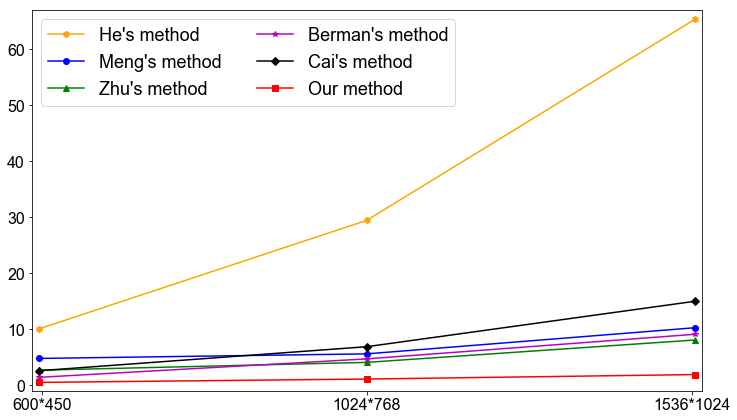}
\caption{Speed variation as the image size increases.}\label{fig:cc}
\end{center}
\end{figure}

We observed that the processing time monotonically increases as the image size grows. To compare the model scalability, we investigate the slope as in Fig.~\ref{fig:cc}. We interpolate Cai's learning-based method~\cite{cai16} as reported in the literature, while running time of other methods are obtained through experiments. It is shown that Zhu's method~\cite{Zhu2015} and Berman's method ~\cite{Dana2016} has roughly a $\mathcal{O}(n^2)$ complexity to the image size, or $\mathcal{O}(n)$ complexity to the number of pixels. Other methods, especially by He~\cite{Guided} have much higher computational complexity than this. Our model has a \emph{quasi-linear} complexity to the image size, that is, $\mathcal{O}(n)$ to the image size, or $\mathcal{O}(\sqrt{n})$ complexity to the number of pixels. We believe this exceptional speed should be attributed to the significant dimension reduction effect of a 2-dimensional DHWT, which is considered highly efficient providing the satisfying dehazing quality.

\section{Conclusion}\label{sec:conclusion}
Convex optimization based on the optical model of image can efficiently perform high-quality image dehazing. The method has several advantages, e.g.,~consistent with the physical theory of haze formation and the inverse problem can be solved at the same time. Due to these reasons, the method is less suffered from over-saturation and halos, which are common among other dehazing algorithms. In this paper, we introduce multilevel wavelet transform to reduce the dimension of the original image, and perform optimization based image dehazing on the low frequency sub-band. Finally, the dehazed image is recovered using inverse discrete wavelet transform. This strategy produces the dehazing results comparable to the state-of-the-art methods. Meanwhile, the computational complexity of the proposed method is reduced to quasi-linear to the image size. Experiments from various aspects support the aforementioned findings. In future, we will study new loss items and parameter learning, such as the optimal level of wavelet transform and regularizer weighting.

\section{Acknowledgement}
The authors thank Jean-Philippe Tarel, Kaiming He, Qingsong Zhu, Gaofeng Meng, Dana Berman, Yuankai Wang, Raanan Fattal, and Ko Nishino for providing their algorithms and experimental results.
\newpage
\noindent\textbf{\large{References}}
\bibliography{fsid}

\begin{thebibliography}{10}
\expandafter\ifx\csname url\endcsname\relax
  \def\url#1{\texttt{#1}}\fi
\expandafter\ifx\csname urlprefix\endcsname\relax\def\urlprefix{URL }\fi
\expandafter\ifx\csname href\endcsname\relax
  \def\href#1#2{#2} \def\path#1{#1}\fi

\bibitem{rosa17}
R.~Azami, D.~Mould, Detail and color enhancement in photo stylization, in:
  Proceedings of the ACM symposium on Computational Aesthetics, 2017.

\bibitem{xing16}
F.~Xing, E.~Cambria, W.-B. Huang, Y.~Xu, Weakly supervised semantic
  segmentation with superpixel embedding, in: IEEE International Conference on
  Image Processing (ICIP), 2016, pp. 1269--1273.

\bibitem{christos18}
C.~Sakaridis, D.~Dai, S.~Hecker, L.~V. Gool, Model adaptation with synthetic
  and real data for semantic dense foggy scene understanding, in: The European
  Conference on Computer Vision (ECCV), 2018, pp. 687--704.

\bibitem{goyal18}
K.~Goyal, J.~Singhai, Review of background subtraction methods using gaussian
  mixture model for video surveillance systems, Artificial Intelligence Review
  50~(2) (2018) 241--259.

\bibitem{singh-18}
D.~Singh, V.~Kumar, A comprehensive review of computational dehazing
  techniques, Archives of Computational Methods in Engineering (2018)
  1--19\href {http://dx.doi.org/http://doi.org/10.1007/s11831-018-9294-z}
  {\path{doi:http://doi.org/10.1007/s11831-018-9294-z}}.

\bibitem{wangw18}
W.~Wang, F.~Chang, T.~Ji, X.~Wu, A fast single-image dehazing method based on a
  physical model and gray projection, IEEE Access 6 (2018) 5641--5653.

\bibitem{Schechner2006}
S.~Shwartz, E.~Namer, Y.~Y. Schechner, Blind haze separation, in: IEEE
  Conference on Computer Vision and Pattern Recognition (CVPR), Vol.~2, 2006,
  pp. 1984--1991.

\bibitem{Schechner2007}
Y.~Y. Schechner, Y.~Averbuch, Regularized image recovery in scattering media,
  IEEE Transactions on Pattern Analysis and Machine Intelligence (TPAMI) 29~(9)
  (2007) 1655--1660.

\bibitem{Hautiere2007}
N.~Hautiere, J.~Tarel, D.~Aubert, Toward fog-free in-vehicle vision systems
  through contrast restoration, in: IEEE Conference on Computer Vision and
  Pattern Recognition (CVPR), 2007.

\bibitem{Kopf2008}
J.~Kopf, B.~Neubert, B.~Chen, M.~Cohen, D.~Cohen-Or, O.Deussen, M.Uyttendaele,
  D.~Lischinski, Deep photo: Model-based photograph enhancement and viewing,
  in: ACM Transactions on Graphics, Vol.~27, 2008, pp. 1--10.

\bibitem{luan17}
Z.~Luan, Y.~Shang, X.~Zhou, Z.~Shao, G.~Guo, X.~Liu, Fast single image dehazing
  based on a regression model, Neurocomputing 245 (2017) 10--22.

\bibitem{He2011}
K.~He, X.~Tang, Single image haze removal using dark channel prior, IEEE
  transactions on Pattern Analysis and Machine Intelligence (TPAMI) 33~(12)
  (2011) 2341--2353.

\bibitem{Guided}
K.~He, J.~Sun, X.~Tang, Guided image filtering, IEEE Transactions on Pattern
  Analysis and Machine Intelligence (TPAMI) 35~(6) (2012) 1397 -- 1409.

\bibitem{Gao2014}
Y.~Gao, H.-M. Hu, S.~Wang, B.~Li, A fast image dehazing algorithm based on
  negative correction, Signal Processing 103 (2014) 380--398.

\bibitem{Fattal2008}
R.~Fattal, Single image dehazing, ACM Transactions on Graphics 27~(3) (2008)
  1--9.

\bibitem{Bayesian-defog}
K.~Nishino, L.~Kratz, S.~Lmbardi, Bayesian defogging, International Journal of
  Computer Vision 98~(3) (2012) 263--278.

\bibitem{Mutimbu}
L.~Mutimbu, A.~Robles-Kelly, A relaxed factorial markov random field for colour
  and depth estimation from a single foggy image, in: IEEE International
  Conference on Image Processing (ICIP), 2013, pp. 355--359.

\bibitem{Wang2014}
Y.-K. Wang, C.-T. Fan, Single image defogging by multiscale depth fusion, IEEE
  Transactions on Image Processing 23~(11) (2014) 4826--4837.

\bibitem{Tarel2009}
J.-P. Tarel, N.~Hauti{\`e}re, Fast visibility restoration from a single color
  or gray level image, in: IEEE International Conference on Computer Vision
  (ICCV), 2009, pp. 2201--2208.

\bibitem{Ancuti2011}
C.~O. Ancuti, C.~Ancuti, C.~Hermans, P.~Bekaert, A fast semi-inverse approach
  to detect and remove the haze from a single image, in: Asian Conference on
  Computer Vision (ACCV), 2010, pp. 501--514.

\bibitem{Zhang2011}
J.~Zhang, L.~Li, Y.~Zhang, G.~Yang, X.~Cao, J.~Sun, Video dehazing with spatial
  and temporal coherence, The Visual Computer 27~(6) (2011) 749--757.

\bibitem{Kim2013}
J.-H. Kim, W.-D. Jang, J.-Y. Sim, C.-S. Kim, Optimized contrast enhancement for
  real-time image and video dehazing, Journal of Visual Communication and Image
  Representation 24~(3) (2013) 410--425.

\bibitem{Zhu2015}
Q.~Zhu, J.~Mai, L.~Shao, A fast single image haze removal algorithm using color
  attenuation prior, IEEE Transactions on Image Processing 24~(11) (2015)
  3522--3533.

\bibitem{Zhao-mof}
D.~Zhao, L.~Xu, YihuaYan, J.~Chen, L.-Y. Duan, Learning intensity and detail
  mapping parameters for dehazing, Signal Processing: Image Communication 74
  (2019) 253--265.

\bibitem{Yang2012}
Z.~Yang, C.~Zhang, L.~Xie, Robustly stable signal recovery in compressed
  sensing with structured matrix perturbation, IEEE Transactions on Signal
  Processing 60~(9) (2012) 4658--4671.

\bibitem{he16}
J.~He, C.~Zhang, R.~Yang, K.~Zhu, Convex optimization for fast image dehazing,
  in: International Conference on Image Processing (ICIP), 2016, pp.
  2246--2250.

\bibitem{Tarel2012}
J.-P. Tarel, N.~Hauti{\`e}re, L.~Caraffa, A.~Cord, H.~Halmaoui, D.~Gruyer,
  Vision enhancement in homogeneous and heterogeneous fog, IEEE Intelligent
  Transportation Systems Magazine 4~(2) (2012) 6--20.

\bibitem{Narasimhan2002}
S.~G. Narasimhan, S.~K. Nayar, Vision and atmosphere, International Journal of
  Computer Vision 48~(3) (2002) 233--254.

\bibitem{wangw17}
W.~Wang, X.~Yuan, X.~Wu, Y.~Liu, Fast image dehazing method based on linear
  transformation, IEEE Transactions on Multimedia 19~(6) (2017) 1142--1155.

\bibitem{airlight}
C.~Ancuti, C.~O. Ancuti, C.~D. Vleeschouwer, A.~C. Bovik, Night-time dehazing
  by fusion, in: IEEE International Conference on Image Processing (ICIP),
  2016, pp. 2256--2260.

\bibitem{higham2002}
N.~J. Higham, Accuracy and Stability of Numerical Algorithms, SIAM: Society for
  Industrial and Applied Mathematics, 2002.

\bibitem{TV}
L.~I. Rudin, S.~Osher, E.~Fatemi, Nonlinear total variation based noise removal
  algorithms, Physica D: Nonlinear Phenomena 60~(1) (1992) 259--268.

\bibitem{Meng}
G.~Meng, Y.~Wang, J.~Duan, S.~Xiang, C.~Pan, Efficient image dehazing with
  boundary constraint and contextual regularization, in: IEEE International
  Conference on Computer Vision (ICCV), 2013, pp. 617--624.

\bibitem{Dana2016}
D.~Berman, T.~Treibitz, S.~Avidan, Non-local image dehazing, in: The IEEE
  Conference on Computer Vision and Pattern Recognition (CVPR), 2016, pp.
  1674--1682.

\bibitem{SplitBregman}
T.~Goldstein, S.~Osher, The split bregman method for l1-regularized problems,
  SIAM Journal on Imaging Sciences 2~(2) (2009) 323--343.

\bibitem{ihaze}
C.~O. Ancuti, C.~Ancuti, R.~Timofte, C.~D. Vleeschouwer, I-haze: a dehazing
  benchmark with real hazy and haze-free indoor images (2018).
\newblock \href {http://arxiv.org/abs/arXiv:1804.05091}
  {\path{arXiv:arXiv:1804.05091}}.

\bibitem{ohaze}
C.~O. Ancuti, C.~Ancuti, R.~Timofte, C.~D. Vleeschouwer, O-haze: a dehazing
  benchmark with real hazy and haze-free outdoor images (2018).
\newblock \href {http://arxiv.org/abs/arXiv:1804.05101}
  {\path{arXiv:arXiv:1804.05101}}.

\bibitem{Tarel2008}
N.~Hautiere, J.-P. Tarel, J.~Lavenant, D.~Aubert, Blind contrast enhancement
  assessment by gradient ratioing at visible edges, Image Analysis \&
  Stereology Journal 27 (2008) 87--95.

\bibitem{zwang04}
Z.~Wang, A.~Bovik, H.~Sheikh, E.~Simoncelli, Image quality assessment: from
  error visibility to structural similarity, IEEE Transactions on Image
  Processing 13~(4) (2004) 600--612.

\bibitem{mngpp}
D.~Singh, V.~Kumar, Image dehazing using moore neighborhood-based gradient
  profile prior, Signal Processing: Image Communication 70 (2019) 131--144.

\bibitem{Lian-18}
X.~Lian, Y.~Pang, A.~Yang, Learning intensity and detail mapping parameters for
  dehazing, Multimedia Tools and Applications 77~(12) (2018) 15695--15720.

\bibitem{cai16}
B.~Cai, X.~Xu, K.~Jia, C.~Qing, D.~Tao, Dehazenet: An end-to-end system for
  single image haze removal, IEEE Transaction on Image Processing 25~(11)
  (2016) 5187--5198.

\end{thebibliography}
\end{document}